\documentclass{article}

\usepackage{arxiv}
\usepackage[utf8]{inputenc} 
\usepackage[T1]{fontenc}    
\usepackage{hyperref}       
\usepackage{url}            
\usepackage{booktabs}       
\usepackage{amsfonts}       
\usepackage{nicefrac}       
\usepackage{microtype}      
\usepackage{cool}

\usepackage{url,hyperref,lineno,microtype}
\usepackage{mathtools}
\usepackage{amsmath}
\usepackage{algorithm}
\usepackage[noend]{algpseudocode}
\usepackage{soul}
\usepackage{physics}
\usepackage{multicol}
\usepackage{multirow}
\usepackage{times}
\usepackage{textcomp}
\usepackage[caption=false]{subfig}
\usepackage{float}
\usepackage{graphicx}

\usepackage{nameref}

\title{Enabling Spike-based Backpropagation for Training Deep Neural Network Architectures}

\author{
  Chankyu Lee\thanks{These two authors contributed equally.} \\
  Department of Electrical and Computer Engineering\\
  Purdue University\\
  West Lafayette, IN, 47907 \\
  \texttt{lee2216@purdue.edu} \\
   \And
 Syed Shakib Sarwar$^{*}$ \\
  Department of Electrical and Computer Engineering\\
  Purdue University\\
  West Lafayette, IN, 47907 \\
  \texttt{sarwar@purdue.edu} \\
   \And
    Priyadarshini Panda \\
  Department of Electrical and Computer Engineering\\
  Purdue University\\
  West Lafayette, IN, 47907 \\
  \texttt{pandap@purdue.edu} \\
   \And
    Gopalakrishnan Srinivasan \\
  Department of Electrical and Computer Engineering\\
  Purdue University\\
  West Lafayette, IN, 47907 \\
  \texttt{srinivg@purdue.edu} \\
   \And
 Kaushik Roy \\
  Department of Electrical and Computer Engineering\\
  Purdue University\\
  West Lafayette, IN, 47907 \\
  \texttt{kaushik@purdue.edu} \\
}

\begin{document}
\maketitle

\begin{abstract}
Spiking Neural Networks (SNNs) have recently emerged as a prominent neural computing paradigm. However, the typical shallow SNN architectures have limited capacity for expressing complex representations while training deep SNNs using input spikes has not been successful so far. Diverse methods have been proposed to get around this issue such as converting off-the-shelf trained deep Artificial Neural Networks (ANNs) to SNNs. However, the ANN-SNN conversion scheme fails to capture the temporal dynamics of a spiking system. On the other hand, it is still a difficult problem to directly train deep SNNs using input spike events due to the discontinuous, non-differentiable nature of the spike generation function. To overcome this problem, we propose an approximate derivative method that accounts for the leaky behavior of LIF neurons. This method enables training deep convolutional SNNs directly (with input spike events) using spike-based backpropagation. Our experiments show the effectiveness of the proposed spike-based learning on deep networks (VGG and Residual architectures) by achieving the best classification accuracies in MNIST, SVHN and CIFAR-10 datasets compared to other SNNs trained with a spike-based learning. Moreover, we analyze sparse event-based computations to demonstrate the efficacy of the proposed SNN training method for inference operation in the spiking domain.
\end{abstract}


\section{Introduction}
Over the last few years, deep learning has made tremendous progress and has become a prevalent tool for performing various cognitive tasks such as object detection, speech recognition and reasoning. Various deep learning techniques \cite{lecun1998gradient,srivastava2014dropout,ioffe2015batch} enable the effective optimization of deep ANNs by constructing multiple levels of feature hierarchies and show remarkable results, which occasionally outperform human-level performance \cite{krizhevsky2012imagenet, he2016deep, silver2016mastering}. To harness the deep learning capabilities in ubiquitous environments, it is necessary to deploy deep learning not only on large-scale computers, but also on edge devices  (\textit{e.g.} phone, tablet, smartwatch, robot, etc.). However, the ever-growing complexity of deep neural networks together with the explosion in the amount of data to be processed, place significant energy demands on current computing platforms.

Spiking Neural Network (SNN) is one of the leading candidates for overcoming the constraints of neural computing and to efficiently harness the machine learning algorithm in real-life (or mobile) applications. The concepts of SNN, which is often regarded as the 3\textsuperscript{rd} generation neural network \cite{maass1997networks}, are inspired by the biological neuronal mechanisms \cite{dayan2001theoretical, hodgkin1952currents,brette2005adaptive,izhikevich2003simple} that can efficiently process discrete spatio-temporal events (spikes). The Leaky Integrate and Fire (LIF) neuron is the simple first-order phenomenological spiking neuron model, which can be characterized by the internal state, called membrane potential. The membrane potential integrates the inputs over time and generates an output spike whenever it overcomes the neuronal firing threshold. Recently, specialized hardwares \cite{merolla2014million,ankit2017resparc, davies2018loihi} have been developed to exploit this event-based, asynchronous signaling/processing scheme. They are promising for achieving ultra-low power intelligent processing of streaming spatiotemporal data, and especially in deep hierarchical networks, as it has been observed in SNN models that the number of spikes, and thus the amount of computation, decreases significantly at deeper layers \cite{sengupta2018going,rueckauer2017conversion}.

We can divide SNNs into two broad classes: a) converted SNNs and b) SNNs derived by direct spike-based training. The former one is SNNs converted from the trained ANN for the efficient event-based inference (ANN-SNN conversion) \cite{cao2015spiking,diehl2016conversion,hunsberger2015spiking,sengupta2018going,rueckauer2017conversion}. The main advantage is that it uses state-of-the-art (SOTA), optimization-based, ANN training methods and therefore achieves SOTA classification performance. For instance, the specialized SNN hardwares (such as SpiNNaker \cite{furber2013overview}, IBM TrueNorth \cite{merolla2014million}) have exhibited greatly improved power efficiency as well as the state-of-the-art performance for the inference. The signals used in such training are real-valued and are naturally viewable as representing spike rate (frequency). The problem is that reliably estimating frequencies requires nontrivial passage of time. On the other hand, SNNs derived by direct spike-based training also involves some hurdles. Direct spike-based training methods can be divided into two classes: i) non-optimization-based, mostly unsupervised, approaches involving only signals local to the synapse, e.g., the times of pre- and post-synaptic spikes, as in Spike-Timing-Dependent-Plasticity (STDP); and ii) optimization-based, mostly supervised, approaches involving a global objective, e.g., loss, function. STDP-trained two-layer network (consisting of 6400 output neurons) has been shown to achieve 95\% classification accuracy on MNIST dataset \cite{diehl2015unsupervised}.

However, a shallow network structure limits the expressive power \cite{zhao2015feedforward,brader2007learning,srinivasan2018spilinc,srinivasan2018stdp} and may not scale well to larger problem sizes. While efficient feature extraction has been demonstrated using layer-wise STDP learning in deep convolutional SNNs \cite{kheradpisheh2016stdp,lee2018deep}, ANN models trained with standard backpropagation (BP) \cite{rumelhart1985learning} still achieve significantly better classification performance. These considerations have inspired the search for spike-based versions of BP, which requires finding a differentiable surrogate of the spiking unit's activation function. \cite{bohte2002error,lee2016training} defined such a surrogate in terms of a unit's membrane potential. Besides, \cite{panda2016unsupervised} applies BP-based supervised training for the classifier after layer-by-layer autoencoder-based training of the feature extractor. By combining layer-wise STDP-based unsupervised and supervised spike-based BP, \cite{lee2018training} showed improved robustness, generalization ability, and faster convergence. In this paper, we take these prior works forward to effectively train very deep SNNs using end-to-end spike-based BP learning.

The main contributions of our work are as follows. First, we develop a spike-based supervised gradient descent BP algorithm that employs an approximate (pseudo) derivative for LIF neuronal function. In addition, we leverage the key idea of the successful deep ANN models such as LeNet5 \cite{lecun1998gradient}, VGG \cite{simonyan2014very} and ResNet \cite{he2016deep} for efficiently constructing deep convolutional SNN architectures. We also adapt the dropout \cite{srivastava2014dropout} technique to better regularize deep SNN training. Next, we demonstrate the effectiveness of our methodology for visual recognition tasks on standard character and object datasets (MNIST, SVHN, CIFAR-10) and a neuromorphic dataset (N-MNIST). To the best of our knowledge, this work achieves the best classification accuracy in MNIST, SVHN, and CIFAR-10 datasets among other spike-based learning methodologies. Last, we expand our efforts to quantify and analyze the advantages of a spike-based BP algorithm compared to ANN-SNN conversion techniques in terms of inference time and energy consumption.

The rest of the paper is organized as follows. In section 2, we provide the background on fundamental components and architectures of deep convolutional SNNs. In section 3.1, we detail the spike-based gradient descent BP learning algorithm. In section 3.2, we describe our spiking version of the dropout technique. In section 4, we describe the experiments and report the simulation results, which validate the efficacy of spike-based BP training for MNIST, SVHN, CIFAR-10 and N-MNIST datasets. In section 5.1, we discuss the proposed algorithm in comparison to relevant works. In section 5.2-5.5, we analyze the spike activity, inference speedup and complexity reduction of directly trained SNNs and ANN-SNN converted networks. Finally, we summarize and conclude the paper in section 6.

\begin{figure*}[h]
\centering
\includegraphics[width=\textwidth]{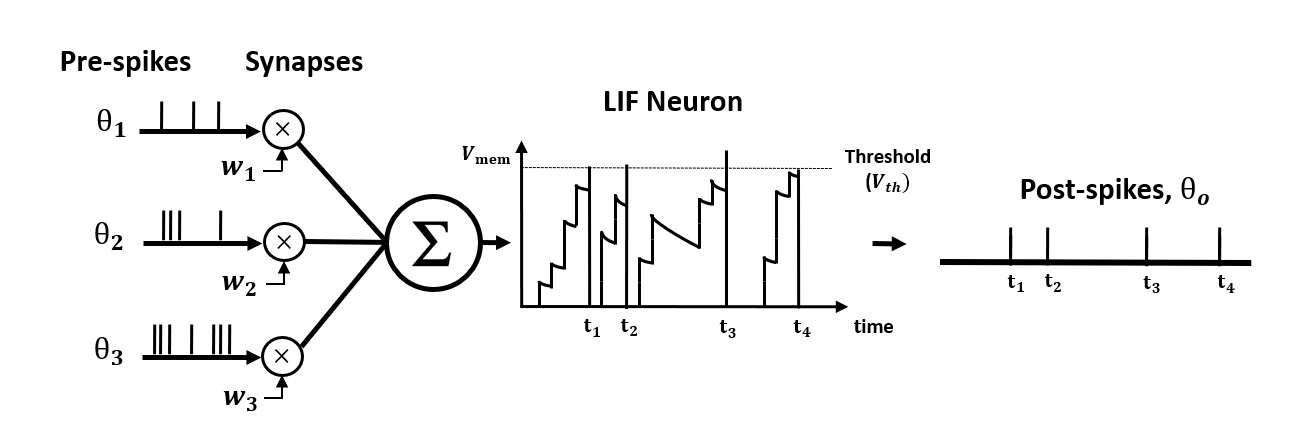}
\caption{The illustration of Leaky Integrate and Fire (LIF) neuron dynamics. The pre-spikes are modulated by the synaptic weight to be integrated as the current influx in the membrane potential that decays exponentially. Whenever the membrane potential crosses the firing threshold, the post-neuron fires a post-spike and resets the membrane potential.}
\label{fig:fig1}
\end{figure*}

\section{The Component and Architecture of Spiking Neural Networks}
\subsection{Spiking Neural Network Component}
Leaky-Integrate-and-Fire (LIF) neurons \cite{dayan2001theoretical} and plastic synapses are fundamental and biologically plausible computational elements for emulating the dynamics of SNNs. The sub-threshold dynamics of a LIF spiking neuron can be formulated as
\[\tau_{m} \frac{dV_{mem}}{dt}= -V_{mem} + I(t) \tag{1}\]
where $V_{mem}$ is the post-neuronal membrane potential and $\tau_{m}$ is the time constant for membrane potential decay. The input current, $I(t)$, is defined as the weighted summation of pre-spikes at each time step as given below.
\[I(t) = \sum_{i=1}^{n^{l}} (w_{i} \sum_{k}\theta_{i} (t-t_{k})) \tag{2}\]
where $n^{l}$ indicates the number of pre-synaptic weights, $w_{i}$ is the synaptic weight connecting $i^{th}$ pre-neuron to post-neuron. $\theta_{i}(t-t_k)$ is a spike event from $i^{th}$ pre-neuron at time $t_k$, which can be formulated as a Kronecker delta function as follows,
\[\theta(t-t_{k}) =\begin{cases}
               1, & \text{if~$t=t_k$}\\
               0, & \text{otherwise}
            \end{cases} \tag{3}\]

\begin{table}[h]
\centering
\caption{List of notations}
\label{table:notation}
\begin{tabular}{|l|l|}
\hline
\textbf{Notations}                                                                            & \textbf{Meaning}            \\ \hline
\begin{tabular}[c]{@{}l@{}} $\theta$ \end{tabular}                   & \begin{tabular}[c]{@{}l@{}} Spike event \end{tabular}    \\ \hline
$x$                                                                       & The sum of pre-spike events over time                  \\ \hline
\begin{tabular}[c]{@{}l@{}}$w$\end{tabular}                                     & Synaptic weight                 \\ \hline
\begin{tabular}[c]{@{}l@{}}$V_{mem}$ \end{tabular}                & Membrane potential        \\ \hline
\begin{tabular}[c]{@{}l@{}}$V_{th}$ \end{tabular}                & Neuronal firing threshold        \\ \hline
$I$                                                      & Input current at each time step              \\ \hline
$net$                                                      & Total current influx over time              \\ \hline
$a$                                               & Activation of spiking neuron                         \\ \hline
$E$                                               & Loss function                         \\ \hline
$\delta$                                              & Error gradient                         \\ \hline
\end{tabular}
\end{table}
            
where $t_k$ is the time instant that $k^{th}$ spike occurred. Figure \ref{fig:fig1} illustrates LIF neuronal dynamics. The impact of each pre-spike, $\theta_{i}(t-t_{k})$, is modulated by the corresponding synaptic weight ($w_{i}$) to generate a current influx to the post-neuron. Note, the units do not have bias term. The input current is integrated into the post-neuronal membrane potential ($V_{mem}$) that leaks exponentially over time with time constant ($\tau_{m}$). When the membrane potential exceeds a threshold ($V_{th}$), the neuron generates a spike and resets its membrane potential to initial value. The table \ref{table:notation} lists the annotations used in equations (1-27).

\subsection{Deep Convolutional Spiking Neural Network}
\subsubsection{Building Blocks}
\begin{figure*}[h]
\centering
\includegraphics[width=\textwidth]{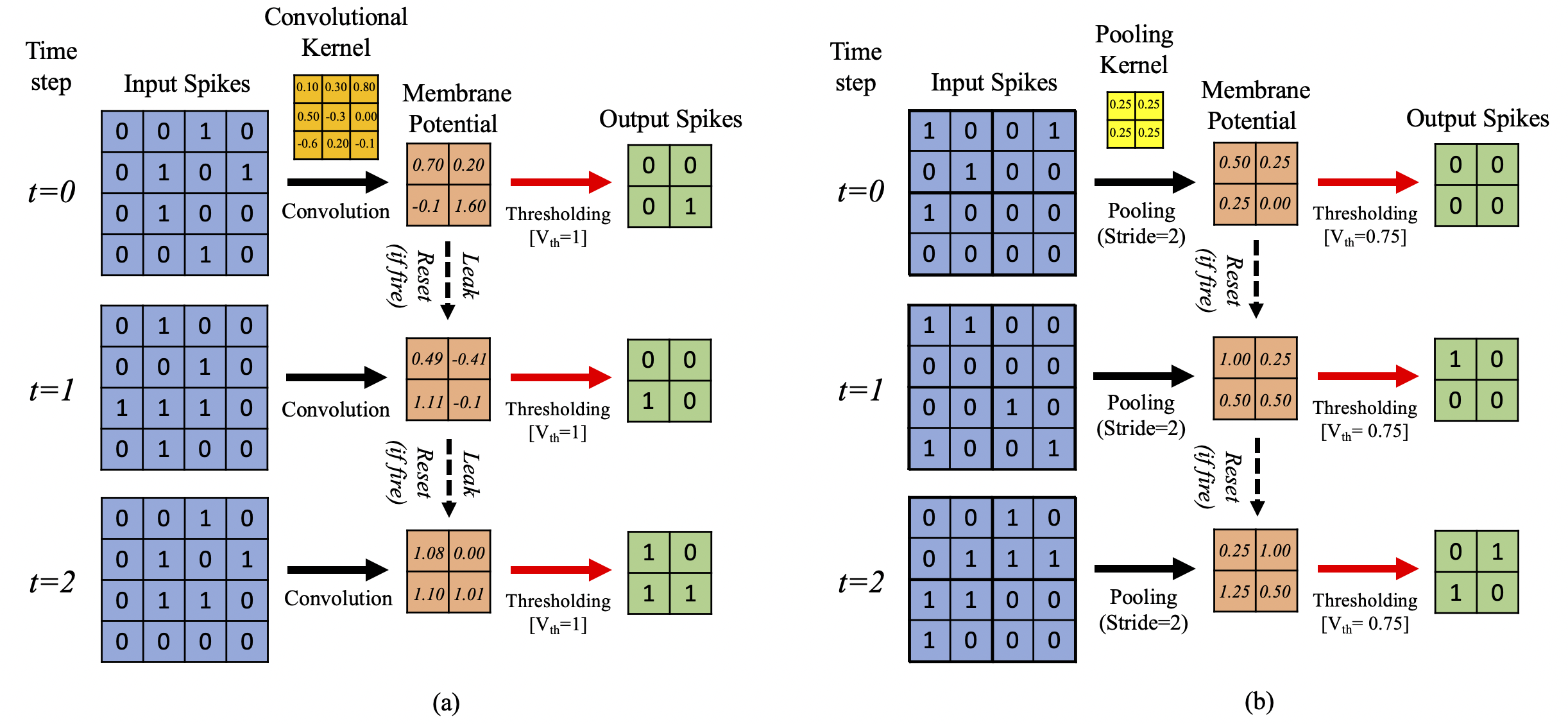}
\caption{Illustration of the simplified operational example of (a) convolutional, (b) spatial-pooling layers (assuming 2-D input and 2-D weight kernel) over three time steps.
At each time step, the input spikes are convolved with the weight kernel to generate the current influx, which is accumulated in the post-neuron's membrane potential, $V_{mem}$. Whenever the membrane potential exceeds the firing threshold ($V_{th}$), the post-neuron in the output feature map spikes and $V_{mem}$ resets. Otherwise, $V_{mem}$ is considered as residue in the next time step while leaking in the current time step. For spatial-pooling, the kernel weights are fixed, and there is no membrane potential leak.}
\label{fig:fig21}
\end{figure*}

In this work, we develop a training methodology for convolutional SNN models that consist of an input layer followed by intermediate hidden layers and a final classification layer. In the input layer, the pixel images are encoded as Poisson-distributed spike trains where the probability of spike generation is proportional to the pixel intensity. The hidden layers consist of multiple convolutional (C) and spatial-pooling (P) layers, which are often arranged in an alternating manner. These convolutional (C) and spatial-pooling (P) layers represent the intermediate stages of feature extractor. The spikes from the feature extractor are combined to generate a one-dimensional vector input for the fully-connected (FC) layers to produce the final classification. The convolutional and fully-connected layers contain trainable parameters (\textit{i.e.} synaptic weights) while the spatial-pooling layers are fixed \textit{a priori}. Through the training procedure, weight kernels in the convolutional layers can encode the feature representations of the input patterns at multiple hierarchical levels. Figure \ref{fig:fig21}a shows the simplified operational example of a convolutional layer consisting of LIF neurons over three time steps (assuming 2-D input and 2-D weight kernel). On each time step, each neuron convolves its input spikes with the weight kernel to compute its input current, which is integrated into its membrane potential, $V_{mem}$. If $V_{mem} > V_{th}$, the neuron spikes and its $V_{mem}$ is set to 0. Otherwise, $V_{mem}$ is considered as residue in the next time step while leaking in the current time step. Figure \ref{fig:fig21}b shows the simplified operation of a pooling layer, which reduces the dimensionality from the previous convolutional layer while retaining spatial (topological) information.\par

There are various choices for performing the spatial-pooling operation in the ANN domain. The The two major operations used for pooling are max and average. 
Both have been used for SNNs, e.g., max-pooling \cite{rueckauer2017conversion} and average-pooling\cite{diehl2015fast,cao2015spiking}. We use average-pooling due to its simplicity. In the case of SNNs, an additional thresholding is used after averaging to generate output spikes. For instance, a fixed 2$\times$2 kernel (each having a weight of 0.25) strides through a convolutional feature map without overlapping and fires an output spike at the corresponding location in the pooled feature map only if the sum of the weighted spikes of the 4 inputs within the kernel window exceeds a designated firing threshold (set to 0.75 in this work). Otherwise, the membrane potential remains as a residue in the next time step. Figure \ref{fig:fig21}b shows an example spatial-pooling operation over three time steps (assuming 2-D input and 2-D weight kernel). The average-pooling threshold need to be carefully set so that spike propagation is not disrupted due to the pooling. If the threshold is too low, there will be too many spikes, which can cause loss of spatial location of the feature that was extracted from the previous layer. If the threshold is too high, there will not be enough spike propagation to the deeper layers.

\begin{figure*}[h]
\centering
\subfloat[]{\label{fig2a}
  \centering
  \includegraphics[width=.415\textwidth]{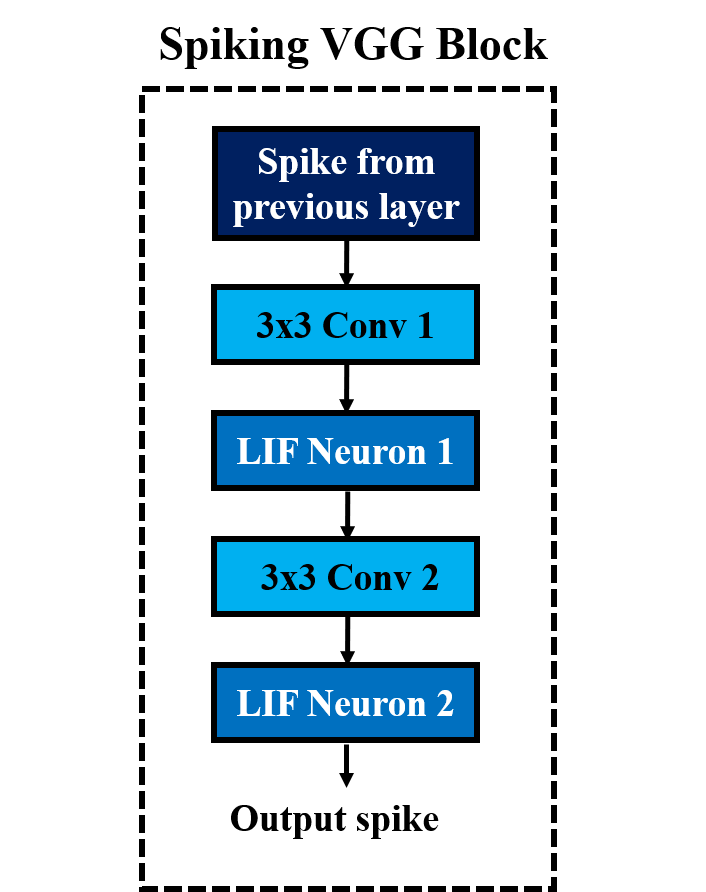}}
\subfloat[]{\label{fig2b}
  \centering
  \includegraphics[width=.5\textwidth]{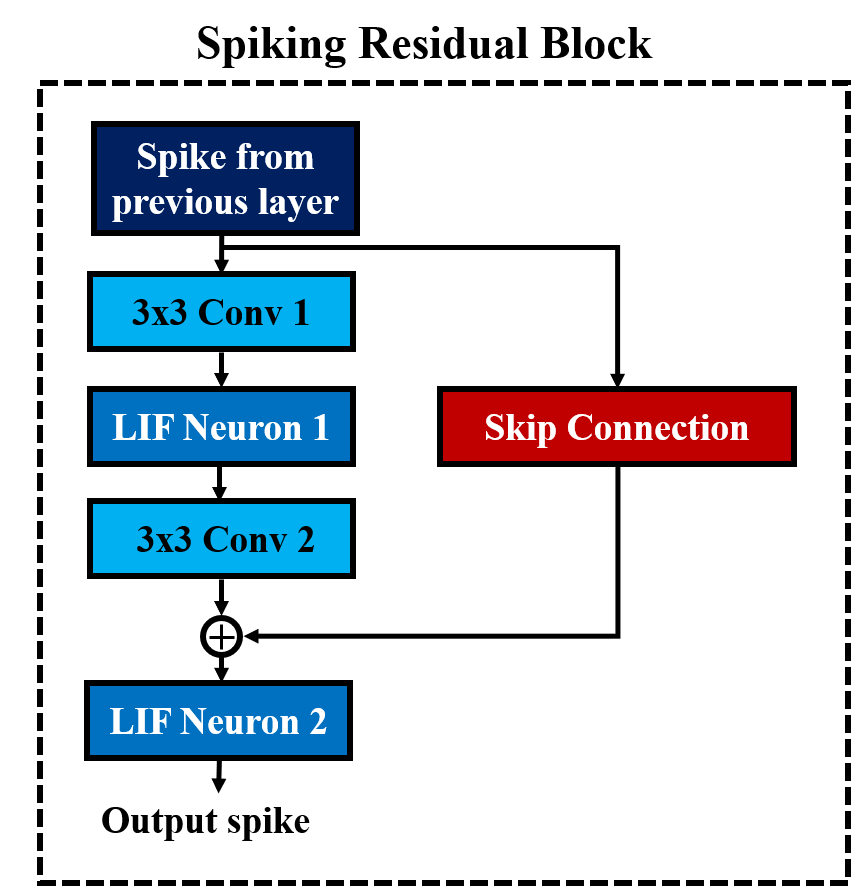}}
\caption{The basic building blocks of the described convolutional SNN architectures. (a) Spiking VGG Block. (b) Spiking ResNet Block.} 
\label{fig:fig2}
\end{figure*}

\subsubsection{Deep Convolutional SNN architecture: VGG and Residual SNNs}
Deep networks are essential for recognizing intricate input patterns so that they can effectively learn hierarchical representations. To that effect, we investigate popular deep neural network architectures such as VGG \cite{simonyan2014very} and ResNet \cite{he2016deep} in order to build deep SNN architectures. VGG \cite{simonyan2014very} was one of the first neural networks, which used the idea of using small (3$\times$3) convolutional kernels uniformly throughout the network. Using small kernels enables effective stacking of convolutional layers while minimizing the number of parameters in deep networks. In this work, we build deep convolutional SNNs (containing more than 5 trainable layers) using `Spiking VGG Block's, which contain stacks of convolutional layers using small (3$\times$3) kernels. Figure \ref{fig2a} shows a `Spiking VGG block' containing two stacked convolutional layers, each followed by a LIF neuronal layer. The convolutional layer box contains the synaptic connectivity, and the LIF neuronal box contains the activation units. Next, ResNet \cite{he2016deep} introduced the skip connections throughout the network that had considerable successes in enabling successful training of significantly deeper networks. In particular, ResNet addresses the degradation (of training accuracy) problem \cite{he2016deep} that occurs while increasing the number of layers in the standard feedforward neural network. We employ the concept of skip connection to construct deep residual SNNs with 7-11 trainable layers. Figure \ref{fig2b} shows a `Spiking Residual Block' containing non-residual and residual paths. The non-residual path consists of two convolutional layers with an intermediate LIF neuronal layer. The residual path (skip connection) is composed of the identity mapping when the number of input and output feature maps are the same, and 1$\times$1 convolutional kernels when the number of input and output feature maps differ. The outputs of both the non-residual and residual paths are integrated to the membrane potential in the last LIF neuronal activation layer (LIF Neuron 2 in figure \ref{fig2b}) to generate output spikes from the `Spiking Residual Block'. Within the feature extractor, a `Spiking VGG Block' or `Spiking Residual Block' is often followed by an average-pooling layer. Note, in some `Spiking Residual Blocks', the last convolutional and residual connections employ convolution with a stride of 2 to incorporate the functionality of the spatial-pooling layers. At the end of the feature extractor, extracted features from the last average-pooling layer are fed to a fully-connected layer as a 1-D vector input for inference.


\section{Supervised Training of Deep Spiking Neural Network}
\subsection{Spike-based Gradient Descent Backpropagation Algorithm}
\label{Event-driven BP}
The spike-based BP algorithm in SNN is adapted from standard BP \cite{rumelhart1985learning} in the ANN domain. In standard BP, the network parameters are iteratively updated in a direction to minimize the difference between the final outputs of the network and target labels. The standard BP algorithm achieves this goal by back-propagating the output error through the hidden layers using gradient descent. However, the major difference between ANNs and SNNs is the dynamics of neuronal output. An artificial neuron (such as \textit{sigmoid}, \textit{tanh}, or \textit{ReLU}) communicates via continuous values whereas a spiking neuron generates binary spike outputs over time. In SNNs, spatiotemporal spike trains are fed to the network as inputs. Accordingly, the outputs of spiking neuron are spike events, which are also discrete over time. Hence, the standard BP algorithm is incompatible with training SNNs, as it can not back-propagate the gradient through a non-differentiable spike generation function. In this work, we formulate an approximate derivative for LIF neuron activation, making gradient descent possible. We derive a spike-based BP algorithm that is capable of learning spatiotemporal patterns in spike-trains. The spike-based BP can be divided into three phases, forward propagation, backward propagation and weight update, which we describe in the following sections.\par

\subsubsection{Forward Propagation}
In forward propagation, spike trains representing input patterns are presented to the network for estimating the network outputs. To generate the spike inputs, the input pixel values are converted to Poisson-distributed spike trains and delivered to the network. The input spikes are multiplied with synaptic weights to produce an input current that accumulates in the membrane potential of post neurons as in equation (1). Whenever its membrane potential exceeds a neuronal firing threshold, the post-neuron generates an output spike and resets. Otherwise, the membrane potential decays exponentially over time. The neurons of every layer (excluding output layer) carry out this process successively based on the weighted spikes received from the preceding layer. Over time, the total weighted summation of the pre-spike trains (i.e., $net$) is described as follows,

\[net_{j}^{l}(t) = \sum_{i=1}^{n^{l-1}} (w_{ij}^{l-1} x_{i}^{l-1}(t)),~where~x_{i}^{l-1}(t) = \sum_{t} \sum_{k} \theta_{i}^{l-1} (t-t_{k}) \tag{4}\]

where $net_{j}^{l}(t)$ represents the total current influx integrated to the membrane potential of $j^{th}$ post-neuron in layer $l$ over the time $t$, $n^{l-1}$ is the number of pre-neurons in layer $l$-$1$ and $x^{l-1}_{i}(t)$ denotes the sum of spike train ($t_k \leq t$) from $i^{th}$ pre-neuron over time $t$. The sum of post-spike trains ($t_k \leq t$) is represented by $a^{l}_{j}(t)$ for the $j^{th}$ post-neuron.

\[a_{j}^{l}(t) = \sum_{t} \sum_{k} \theta_{j}^{l} (t-t_{k}) \tag{5}\]

Clearly, the sum of post-spike train ($a^{l}(t)$) is equivalent to the sum of pre-spike train ($x^{l}(t)$) for the next layer. On the other hand, the neuronal firing threshold of the final classification layer is set to a very high value so that final output neurons do not spike. In the final layer, the weighted pre-spikes are accumulated in the membrane potential while decaying over time. At the last time step, the accumulated membrane potential is divided by the number of total time steps (T) in order to quantify the output distribution (\textit{output}) as presented by equation (6).

\[output = \frac{V_{mem}^{L}(T)}{number~of~timesteps}~\tag{6}\]

\begin{figure*}[h]
\centering
\includegraphics[width=1\textwidth]{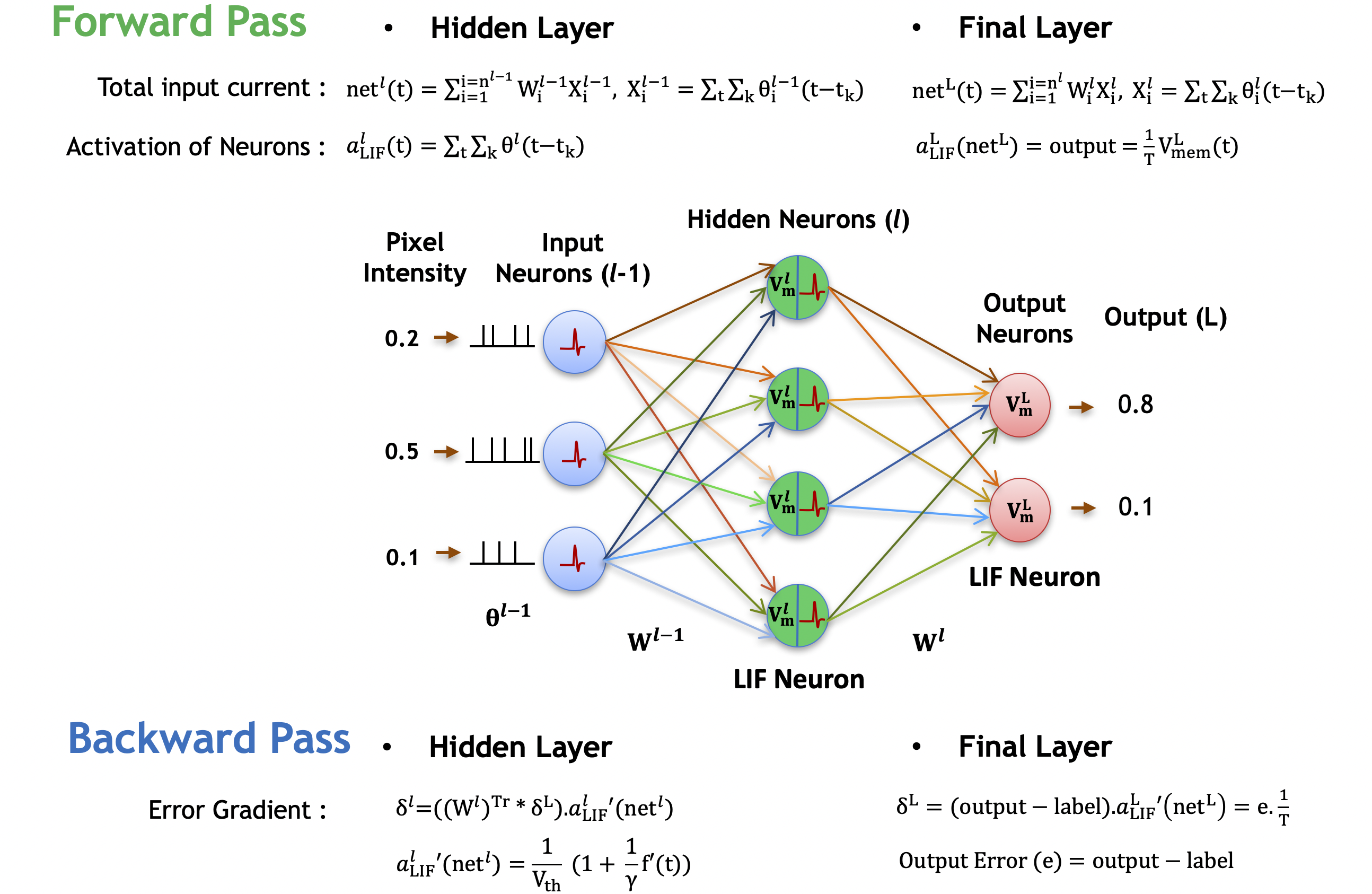}
\caption{Illustration of the forward and backward propagation phase of the proposed spike-based BP algorithm in a multi-layer SNN comprised of LIF neurons. In the forward phase, the LIF neurons (in all layers) accumulate the weighted sum of the pre-spikes in the membrane potential, which decays exponentially over time. In addition, the LIF neurons in hidden layers generate post-spikes if the membrane potential exceeds a threshold and reset the membrane potential. However, the LIF neurons in the final layer, do not generate any spike, but rather accumulate the weighted sum of pre-spikes till the last time step to quantify the final outputs. Then, the final errors are evaluated by comparing the final outputs to the label data. In the backward phase, the final errors are propagated backward through the hidden layers using the chain rule to obtain the partial derivatives of final error with respect to weights. Finally, the synaptic weights are modified in a direction to reduce the final errors.}
\label{fig:fig3}
\end{figure*}

\subsubsection{Backward Propagation and Weight Update}
Next, we describe the backward propagation for the proposed spike-based backpropagation algorithm. After the forward propagation, the loss function is measured as a difference between target labels and outputs predicted by the network. Then, the gradients of the loss function are estimated at the final layer. The gradients are propagated backward all the way down to the input layer through the hidden layers using recursive chain rule, as formulated in equation (7). The following equations (7-27) and figure \ref{fig:fig3} describe the detailed steps for obtaining the partial derivatives of (final) output error with respect to weight parameters.

The prediction error of each output neuron is evaluated by comparing the output distribution (\textit{output}) with the desired target label (\textit{label}) of the presented input spike trains, as shown in equation (8). The corresponding loss function (\textit{E} in equation (9)) is defined as the sum of squared (final prediction) error over all output neurons. To calculate the $\pderiv{E}{a_{LIF}}$ and $\pderiv{a_{LIF}}{net}$ terms in equation (7), we need a defined activation function and 
a method to differentiate the activation function of a LIF neuron.

\[\pderiv{E}{w^l} = \pderiv{E}{a_{LIF}} \pderiv{a_{LIF}}{net} \pderiv{net}{w^l} \tag{7}\]

\[Final~output~error,~e_{j} = output_{j} - label_{j}\tag{8}\]
\[Loss~function,~E = \frac{1}{2} \sum_{j=1}^{n^{L}} {e_{j}}^2 \tag{9}\]

In SNN, the `activation function' indicates the relationship between the weighted summation of pre-spike inputs and post-neuronal outputs over time. In forward propagation, we have different types of neuronal activation for the final layer and hidden layers. Hence, the estimation of neuronal activations and their derivatives are different for the final layer and hidden layers. For the final layer, the value of $output$ in equation (6) is used as the neuronal activation ($a_{LIF}$) while considering the discontinuities at spike time instant as noise. Hence, $\pderiv{E}{output}$ is equal to the final output error, as calculated in equation (10). 
\[\pderiv{E}{output} = \pderiv{~}{output}~\frac{1}{2} (output - label)^2 = output - label = e \tag{10}\]

During back-propagating phase, we consider the leak statistics of membrane potential in the final layer neurons as noise. This allows us to approximate the accumulated membrane potential value for a given neuron as equivalent to the total input current (i.e. $net$) received by the neuron over the forward time duration (T) ($V_{mem,j}^{L}(T) \approx \sum_{i=1}^{n^{L-1}} (w_{ij} x_{i}(T)) = net_{j}^{L}(T)$). Therefore, the derivative of post-neuronal activation with respect to $net$ for final layer ($\pderiv{output}{net} \equiv \pderiv{V_{mem}^{L}(T)/T}{net} = \pderiv{net^{L}(T)/T}{net} = \frac{1}{T}$) is calculated as $\frac{1}{T}$ for the final layer. 


For the hidden layers, we have post-spike trains as the neuronal outputs. The spike generation function is non-differentiable since it creates a discontinuity (because of step jump) at the time instance of firing. Hence, we introduce a pseudo derivative method for LIF neuronal activation ($a_{LIF}'(net)$) for the hidden layers, for back-propagating the output error via the chain rule. The purpose of deriving $a_{LIF}'(net)$ is to approximately estimate the $\pderiv{a_{LIF}}{net}$ term in equation (7) for the hidden layers only. To obtain this pseudo derivative of LIF neuronal activation with respect to total input current (i.e., $net$), we make the following approximations. We first estimate the derivative of an `Integrate and Fire' (IF) neuron's activation. Next, with the derivative of IF neuron's activation, we estimate a leak correctional term to compensate for the leaky effect of membrane potential in LIF activation. Finally, we obtain an approximate derivative for LIF neuronal activation as a combination of two estimations (i.e., derivative for IF neuron and approximated leak compensation derivative). If a hidden neuron does not fire any spike, the derivative of corresponding neuronal activation is set to zero.

\begin{figure*}[h]
\centering
\includegraphics[width=\textwidth]{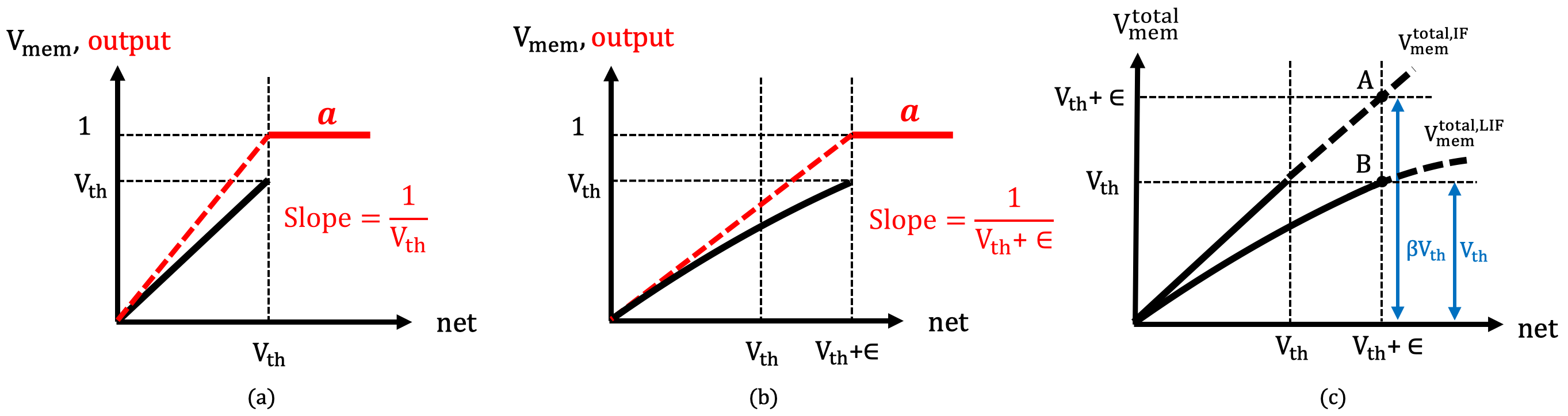}
\caption{(a,b) The illustration of the spike generation function of (a)IF and (b)LIF neuron models, respectively. The x-axis represents the total summation of input currents over time, and y-axis indicates the membrane potential (black) and output (red). The IF neuron generates a post-spike when the input currents accumulated in membrane potential overcome the firing threshold (because of no leaky effect in the membrane potential). However, LIF neuron needs more input currents to cross the firing threshold (because of leaky effect in the membrane potential). Hence, the effective threshold of LIF neurons is considered to be larger compared to the case of IF neurons. (c) The illustration of the estimation of the ratio ($\beta$) between the total membrane potential ($V_{mem}^{total}$) of LIF and IF neurons. If the LIF and IF neuron received the same amount of total input current, the ratio of the total membrane potential of LIF and IF neuron would be estimated as 1:$\beta$ where $\beta$ is greater than 1.}
\label{fig:slope}
\end{figure*}

The spike generation function of IF neuron is a hard threshold function that generates the output signal as either +1 or 0. The IF neuron fires a post-spike whenever the input currents accumulated in membrane potential exceed the firing threshold (note, in case of IF neuron, there is no leak in the membrane potential). Hence, the membrane potential of a post-neuron at time instant $t$ can be written as,

\[V_{mem}(t) \approx \sum_{i=1}^{n} (w_{i} x_{i}(t)) - V_{th} a_{IF}(t) \tag{11}\]

where $n$ denotes the number of pre-neurons, $x_i(t)$ is the sum of spike events from $i^{th}$ pre-neuron over time t (defined in equation (4)) and $a_{IF}(t)$ represents the sum of post-spike trains over time t (defined in equation (5)). In equation (11), $\sum_{i=1}^{n} (w_{i} x_{i}(t))$ accounts for the integration behavior and $V_{th}a_{IF}(t)$ accounts for the fire/reset behavior of the membrane potential dynamics. If we assume $V_{mem}$ as zero (using small signal approximation), the activation of IF neuron ($a_{IF}(t)$) can be formulated as the equation (12). Then, by differentiating it with respect to $net$ (in equation (13)), the derivative of IF neuronal activation can be approximated as a linear function with slope of $\frac{1}{V_{th}}$ as the straight-through estimation \cite{bengio2013estimating}.

\[a_{IF}(t) \approx \frac{1}{V_{th}} \sum_{i=1}^{n} (w_{i} x_{i}(t))  = \frac{1}{V_{th}}net(t) \tag{12}\]
\[\pderiv{a_{IF}}{net} \approx \frac{1}{V_{th}} 1 = \frac{1}{V_{th}} \tag{13}\]

The spike generation function of both the IF and LIF neuron models are the same, namely the hard threshold function. However, the effective neuronal thresholds are considered to be different for the two cases, as shown in figure \ref{fig:slope}a,b. In the LIF neuron model, due to the leaky effect in the membrane potential, larger input current (as compared to IF neuron) needs to be accumulated in order to cross the neuronal threshold and generate a post-spike. Hence, the effective neuronal threshold becomes $V_{th} + \epsilon$ where $\epsilon$ is a positive value that reflects the leaky effect of membrane potential dynamics. Now, the derivative of LIF neuronal activation ($\pderiv{a_{LIF}}{net}$) can be approximated as a hard threshold function (similar to IF and equation (13)) and written as $\frac{1}{V_{th}+\epsilon}$. Clearly, the output of a LIF neuron depends on the firing threshold and leaky characteristics (embodied in $\epsilon$) of the membrane potential whereas the output of an IF neuron depends only on the firing threshold. Next, we explain the detailed steps to estimate the $\epsilon$ and in turn calculate the derivative of LIF neuronal activation ($\pderiv{a_{LIF}}{net}$).


To compute $\epsilon$, the ratio ($\beta$) between the total membrane potential ($V_{mem}^{total}(t)$) of IF and LIF neurons is estimated at the end of forward propagation time (T) as shown in figure \ref{fig:slope}c. Here, $V_{mem}^{total}(t)$ represents the hypothetical total membrane potential with accumulated input current without reset mechanism until time step (t). Suppose both the IF and LIF neurons received the same amount of total input current (i.e. $net(T)$), the total membrane potential of LIF neuron is expected to be lower than the total membrane potential of IF neuron ($V_{mem}^{total,LIF}(T):V_{mem}^{total,IF}(T)=1:\beta$ where $\beta>1$). Hence, by comparing the total membrane potential values of IF and LIF neurons in figure \ref{fig:slope}c, the relation of $\epsilon$ and $\beta$ can be obtained as follows,

\[V_{th}+\epsilon = \beta V_{th} \tag{14}\]

where $V_{th}+\epsilon$ represents the total membrane potential of IF neuron (point A in figure \ref{fig:slope}c) and $V_{th}$ indicates the total membrane potential of LIF neuron (point B in figure \ref{fig:slope}c) when both neurons received the same amount of $net$ inputs. Based on this assumption, we now estimate the ratio ($\beta$) by using the relation of the spike output evolution ($\pderiv{a(t)}{t}$) and the total membrane potential evolution ($\pderiv{V_{mem}^{total}(t)}{t}$) over time as described in equation (16-20). As mentioned previously, the total input current (i.e. $net(t)$) and total membrane potential ($V_{mem}^{total}(t)$) are estimated similar to that of IF neuron (because of no leaky effect) so that  equation (15) can be derived from equation (12). By differentiating equation (15) with respect to time, we get the relation of the spike output evolution ($\pderiv{a_{IF}(t)}{t}$) and the membrane potential evolution ($\pderiv{V_{mem}^{total,IF}(t)}{t}$) over time for IF neuron as described in equation (16).

\[a_{IF}(t) \approx \frac{1}{V_{th}} net(t) \approx \frac{1}{V_{th}} V_{mem}^{total,IF}(t) \tag{15}\]
\[\pderiv{a_{IF}(t)}{t} \approx \frac{1}{V_{th}} \pderiv{V_{mem}^{total,IF}(t)}{t} \tag{16}\]

Hence, in IF neuron case, the evolution of membrane potential over time ($\pderiv{V_{mem}^{total,IF}(t)}{t}$) can be represented by the multiplication of firing threshold ($V_{th}$) and the spike output evolution ($\pderiv{a_{IF}(t)}{t}$) in equation (17). Note, the evolution of membrane potential over time ($\pderiv{V_{mem}^{total,IF}(t)}{t}$) indicates the integration component due to the average input current over time. We consider $a_{IF}(t)$ as homogeneous spike trains where spike firing rates are constant, so that the $\pderiv{a_{IF}(t)}{t}$ can be replaced with the post-neuronal firing rate ($rate(t)$). The homogeneous post-neuronal firing rate, $rate(t)$, can be represented by $\frac{a(t)}{t}$ where $a(t)$ is the number of post-spikes and $t$ means the given forward time window. In LIF neuron case, however, the evolution of membrane potential ($\pderiv{V_{mem}^{total,LIF}(t)}{t}$) can be expressed as the combination of average input current (integration component) and leaky (exponential decay) effect as shown in equation (18). To measure the leaky effect in equation (18), we estimate the low-pass filtered output spikes ($t_k \leq t$) that leak over time using the function $V_{th} f(t)$ (depicted in equation (19)), and differentiate it with respect to time at $t\rightarrow t_k^+$ (from the right-sided limit). The $V_{th} f(t)$, as a post-synaptic potential, contains the total membrane potential history over time. The time constant ($\tau_{m}$) in equation (19) determines the decay rate of post-synaptic potential. Essentially, the main idea is to approximately estimate the leaky effect by comparing the total membrane potential and obtain the ratio ($\beta$) between both cases (i.e. IF and LIF neurons).

\[\pderiv{V_{mem}^{total,IF}(t)}{t} \approx V_{th} \pderiv{a_{IF}(t)}{t} \approx V_{th} rate(t) \tag{17}\]
\[\pderiv{V_{mem}^{total,LIF}(t)}{t} \approx V_{th} rate(t) + V_{th}\pderiv{f(t)}{t} \tag{18}\]
\[f(t) = \sum_{k} exp(-\frac{t-t_{k}}{\tau_{m}}) \tag{19}\]
\[\pderiv{a_{IF}(t)}{t} \approx \frac{1}{V_{th}} \pderiv{V_{mem}^{total,IF}(t)}{t} = \beta \frac{1}{V_{th}} \pderiv{V_{mem}^{total,LIF}(t)}{t} \tag{20}\]

By solving the equation (17, 18, 20), the inverse ratio ($\frac{1}{\beta}$) is derived as follows in equation (21),

\[\frac{1}{\beta} = 1 + \frac{1}{rate(t)} \pderiv{f(t)}{t}  \tag{21}\]

where the first term (unity) indicates the effect of average input currents (that is observed from the approximate derivative of IF neuron activation, namely the straight-through estimation) and the second term ($\frac{1}{rate(t)} \pderiv{f(t)}{t}$) represents the leaky (exponential decay) effect of LIF neuron for the forward propagation time window. Then, by using the relations of $\epsilon$ and $\beta$ in equation (14), the derivative of LIF neuronal activation can be obtained as $\pderiv{a_{LIF}}{net} = \frac{1}{V_{th}+\epsilon} = \frac{1}{\beta V_{th}}$. In this work, to avoid the vanishing gradient phenomena during the error back-propagation, the leaky effect term ($\frac{1}{rate(t)} \pderiv{f(t)}{t}$) is divided by the size of the forward propagation time window (T). Hence, the scaled time derivative of this function, $\frac{1}{\gamma} f’(t)$, is used as the leak correctional term where $\gamma$ denotes the number of output spike events for a particular neuron over the total forward propagation time. As a result, we obtain an approximate derivative for LIF neuronal activation (in hidden layers) as a combination of the straight-through estimation (i.e., approximate derivative of IF neuron activation) and the leak correctional term that compensates leaky effect in the membrane potential as described in equation (22). Please note that, in our work, input and output spikes are not exponentially decaying, the leak only happens according to the mechanism of membrane potential. Moreover, $f(t)$ is not a part of the forward propagation phase, and rather it is only defined to approximately measure the leaky effect during the backward propagation phase by differentiating it with respect to time. The function $f(t)$ is a time-dependent function that simply integrates the output spikes ($t_k \leq t$) temporally, and the resultant sum is decayed over time. It is evident that $f(t)$ is continuous except where spikes occur and the activities jump up \cite{lee2016training}. Therefore, $f(t)$ is differentiable at $t\rightarrow t_k^+$ (from the right-sided limit). Note that, to capture the leaky effect (exponential decay), it is necessary to compute the derivative of $f(t)$ at the points in between the spiking activities, not at the time instant of spiking.

\[\pderiv{a_{LIF}}{net} = \frac{1}{V_{th}+\epsilon} = \frac{1}{\beta V_{th}} \approx \frac{1}{V_{th}} (1+ \frac{1}{\gamma} f'(t)) \\= \frac{1}{V_{th}} (1 + \frac{1}{\gamma} \sum_{k} {-\frac{1}{\tau_{m}}}e^{-\frac{t-t_{k}}{\tau_{m}}})  \tag{22}\]

In summary, the approximations applied to implement a spike-based BP algorithm in SNN are as follows:

\begin{itemize}
  \item During the back-propagating phase, we consider the leaks in the membrane potential of final layer neurons as noises so that the accumulated membrane potential is approximated as equivalent to the total input current ($V_{mem}^{L} \approx net$). Therefore, the derivative of post-neuronal activation with respect to $net$ ($\pderiv{output}{net}$) is calculated as $\frac{1}{T}$ for the final layer.
  \item For hidden layers, we first approximate the activation of an IF neuron as a linear function (i.e., straight-through estimation). Hence, we are able to estimate its derivative of IF neuron's activation \cite{bengio2013estimating} with respect to total input current.
  \item To capture the leaky effect of a LIF neuron (in hidden layers), we estimate the scaled time derivative of the low-pass filtered output spikes that leak over time, using the function $f(t)$. This function is continuous except for the time points where spikes occur \cite{lee2016training}. Hence, it is differentiable in the sections between the spiking activities.
  \item We obtain an approximate derivative for LIF neuronal activation (in hidden layers) as a combination of two derivatives. The first one is the straight-through estimation (i.e., approximate derivative of IF neuron activation). The second one is the leak correctional term that compensates the leaky effect in the membrane potential of LIF neurons. The combination of straight-through estimation and the leak correctional term is expected to be less than 1.
\end{itemize}

Based on these approximations, we can train SNNs with direct spike inputs using a spike-based BP algorithm.

At the final layer, the error gradient, $\delta^{L}$, represents the gradient of the output loss with respect to total input current (i.e., $net$) received by the post-neurons. It can be calculated by multiplying the final output error ($e$) with the derivative of the corresponding post-neuronal activation ($\pderiv{output}{net^L}$) as shown in equation (23). At any hidden layer, the local error gradient, $\delta^{l}$, is recursively estimated by multiplying the back-propagated gradient from the following layer $((w^{l})^{Tr}*\delta^{l+1})$ with derivative of the neuronal activation, $a_{LIF}'(net^{l})$, as presented in equation (24). Note that element-wise multiplication is indicated by ‘.’ while matrix multiplication is represented by ‘*’ in the respective equations.

\[\delta^{L} = \pderiv{E}{output} \pderiv{output}{net^L} = e\frac{1}{T} = \frac{e}{T} \tag{23}\]
\[\delta^{l} = ((w^{l})^{Tr}*\delta^{l+1}).a_{LIF}'(net^{l})\tag{24}\]

The derivative of $net$ with respect to weight is simply the total incoming spikes over time as derived in equation (25). The derivative of the output loss with respect to the weights interconnecting the layers $l$ and $l+1$ ($\triangle w^{l}$ in equation (26)) is determined by multiplying the transposed error gradient at $l+1$ ($\delta^{l+1}$) with the input spikes from layer $l$. Finally, the calculated partial derivatives of loss function are used to update the respective weights using a learning rate ($\eta_{BP}$) as illustrated in equation (27). As a result, iterative updating of the weights over mini-batches of input patterns leads the network state to a local minimum, thereby enabling the network to capture multiple-levels of internal representations of the data.

\[ \pderiv{net}{w^{l}} = \pderiv{~}{w^{l}} (w^{l} * x^{l} (t)) = x^{l} (t) \tag{25}\]
\[ \triangle w^{l} = \pderiv{E}{w^l} = x^{l} (t) * (\delta^{l+1})^{Tr}\tag{26}\]
\[ w^{l}_{updated} = w^{l} - \eta_{BP} \triangle w^{l}  \tag{27}\]

\begin{algorithm*}
 \caption{Forward propagation with dropout at each iteration in SNN}
 \label{Algo1}
 \begin{algorithmic}[1]
    \State \textbf{Input} : Poisson-distributed input spike train ($inputs$), Dropout ratio ($p$), Total number of time steps ($\#timesteps$), Membrane potential ($V_{mem}$), Time constant of membrane potential ($\tau_m$), Firing threshold ($V_{th}$)
        \State \textbf{Initialize} $SNN^{l}.V_{mem} \gets 0~\forall l = 2,...,\#SNN.layer$ 
        \State //~Define the random subset of units (with a probability $1-p$) at each iteration
    \For{$l \leftarrow 1$ to $\#SNN.layer-1$}
        \State $mask^l \gets generate\_random\_subset(probability = 1-p)$  
    \EndFor
    \For{$t \leftarrow 1$ to $\#timesteps$}
        \State //~Set input of first layer equal to spike train of a mini-batch data       
        \State $SNN^{1}.spike[t] \gets inputs[t];$
        \For{$l \leftarrow 2$ to $\#SNN.layer$}         
            \State //~Integrate weighted sum of input spikes to membrane potential
            \State $SNN^{l}.V_{mem}[t] \gets SNN^{l}.V_{mem}[t$-$1] + SNN^{l-1}$$forward(SNN^{l-1}.spike[t]).*(mask^{l-1}/(1$-$p));$
            \State //~If $V_{mem}$ is greater than $V_{th}$, post-neuron generate a spike
        \If {$SNN^{l}.V_{mem}[t] > SNN^{l}.V_{th}$}
        \State //~Membrane potential resets if the corresponding neuron fires a spike
            \State $SNN^{l}.spike[t] \gets 1$
            \State $SNN^{l}.V_{mem}[t] \gets 0$
        \Else
        \State //~Else, membrane potential decays over time
            \State $SNN^{l}.spike[t] \gets 0$
            \State $SNN^{l}.V_{mem}[t] \gets e^{-\frac{1}{\tau_m}} * SNN^{l}.V_{mem}[t]$
        \EndIf
        \EndFor     
    \EndFor
\end{algorithmic}
\end{algorithm*}

\subsection{Dropout in Spiking Nerual Network}
Dropout \cite{srivastava2014dropout} is one of the popular regularization techniques while training deep ANNs. This technique randomly disconnects certain units with a given probability ($p$) to avoid units being overfitted and co-adapted too much to given training data. There are prior works \cite{kappel2015network,kappel2018dynamic,neftci2015unsupervised} that investigated the biological insights on how synaptic stochasticity can provide dropout-like functional benefits in SNNs. In this work, we employ the concept of dropout technique in order to regularize deep SNNs effectively. Note, dropout technique is only applied during training and is not used when evaluating the performance of the network during inference. There is a subtle difference in the way dropout is applied in SNNs compared to ANNs. In ANNs, each epoch of training has several iterations of mini-batches. In each iteration, randomly selected units (with dropout ratio of $p$) are disconnected from the network while weighting by its posterior probability ($\frac{1}{1-p}$). However, in SNNs, each iteration has more than one forward propagation depending on the time length of the spike train. We back-propagate the output error and modify the network parameters only at the last time step. For dropout to be effective in our training method, it has to be ensured that the set of connected units within an iteration of mini-batch data is not changed, such that the neural network is constituted by the same random subset of units during each forward propagation within a single iteration. On the other hand, if the units are randomly connected at each time-step, the effect of dropout will be averaged out over the entire forward propagation time within an iteration. Then, the dropout effect would fade-out once the output error is propagated backward and the parameters are updated at the last time step. Therefore, we need to keep the set of randomly connected units for the entire time window within an iteration. In the experiment, we use the SNN version of dropout technique with the probability ($p$) of omitting units equal to 0.2-0.25. Note that the activations are much sparser in SNN forward propagations compared to ANNs, hence the optimal $p$ for SNNs needs to be less than a typical ANN dropout ratio ($p$=0.5). The details of SNN forward propagation with dropout are specified in Algorithm \ref{Algo1}.

\section{Experimental setup and Result}
\subsection{Experimental Setup}
The primary goal of our experiments is to demonstrate the effectiveness of the proposed spike-based BP training methodology in a variety of deep network architectures. We first describe our experimental setup and baselines. For the experiments, we developed a custom simulation framework using the Pytorch \cite{paszke2017automatic} deep learning package for evaluating our proposed SNN training algorithm. Our deep convolutional SNNs are populated with biologically plausible LIF neurons (with neuronal firing thresholds of unity) in which a pair of pre- and post- neurons are interconnected by plastic synapses. At the beginning, the synaptic weights are initialized with Gaussian random distribution of zero-mean and standard deviation of $\sqrt{\frac{\kappa}{n^l}}$ ($n^l$: number of fan-in~synapses) as introduced in \cite{he2015delving}. Note, the initialization constant $\kappa$ differs by the type of network architecture. For instance, we have used $\kappa=2$  for non-residual network and $\kappa=1$ for residual network. For training, the synaptic weights are trained with a mini-batch spike-based BP algorithm in an end-to-end manner, as explained in section \ref{Event-driven BP}. For static datasets, we train our network models for 150 epochs using mini-batch stochastic gradient descent BP that reduces its learning rate at 70\textsuperscript{th}, 100\textsuperscript{th} and 125\textsuperscript{th} training epochs. For the neuromorphic dataset, we use Adam \cite{kingma2014adam} learning method and reduce its learning rate at 40\textsuperscript{th}, 80\textsuperscript{th} and 120\textsuperscript{th} training epochs. Please, refer to table \ref{table:simparam} for more implementation details. The datasets and network topologies used for benchmarking, the input spike generation scheme for event-based operation and determination of the number of time-steps required for training and inference are described in the following sub-sections.

\begin{table*}[h]
\centering
\caption{Parameters used in the experiments}
\label{table:simparam}
\begin{tabular}{|l|l|}
\hline
\textbf{Parameter}                                                                            & \textbf{Value}            \\ \hline
\begin{tabular}[c]{@{}l@{}}Time Constant of Membrane Potential $(\tau_{m})$\end{tabular}                & 100 time-steps        \\ \hline
\begin{tabular}[c]{@{}l@{}}BP Training Time Duration \end{tabular}                   & \begin{tabular}[c]{@{}l@{}}50-100 time-steps \end{tabular}    \\ \hline
Inference Time Duration                                                                       & Same~as~training                  \\ \hline
\begin{tabular}[c]{@{}l@{}}Mini-batch Size\end{tabular}                                     & 16-32                 \\ \hline
Spatial-pooling Non-overlapping Region/Stride                                                      & 2$\times$2, 2               \\ \hline
Neuronal Firing Threshold                   & 1 (hidden layer), $\infty$ (final layer)                         \\ \hline
Weight~Initialization Constant ($\kappa$)                                               & 2 (non-residual network), 1 (residual network)                         \\ \hline
Learning rate ($\eta_{BP}$)                                               & 0.002 - 0.003                         \\ \hline
Dropout Ratio ($p$)                                               & 0.2 - 0.25                         \\ \hline
\end{tabular}
\end{table*}

\subsubsection{Benchmarking Datasets}
We demonstrate the efficacy of our proposed training methodology for deep convolutional SNNs on three standard vision datasets and one neuromorphic vision dataset, namely the MNIST \cite{lecun1998gradient}, SVHN \cite{netzer2011reading}, CIFAR-10 \cite{krizhevsky2009learning} and N-MNIST \cite{orchard2015converting}. The MNIST dataset is composed of gray-scale (one-dimensional) images of handwritten digits whose sizes are 28 by 28. The SVHN and CIFAR-10  datasets are composed of color (three-dimensional) images whose sizes are 32 by 32. The N-MNIST dataset is a neuromorphic (spiking) dataset that is converted from static MNIST dataset using Dynamic Vision Sensor (DVS)~\cite{lichtsteiner2008128}. The N-MNIST dataset contains two-dimensional images that include ON and OFF event stream data whose sizes are 34 by 34. The ON (OFF) event represents the increase (decrease) in pixel bright changes. The details of the benchmark datasets are listed in table \ref{table:dataset}. For evaluation, we report the top-1 classification accuracy by classifying the test samples (training samples and test samples are mutually exclusive).

\begin{table}[h]
\centering
\caption{Benchmark Datasets}
\label{table:dataset}
\begin{tabular}{|l|l|c|c|c|}
\hline
\textbf{Dataset}   & \textbf{Image}             & \multicolumn{1}{l|}{\textbf{\#Training Samples}} & \multicolumn{1}{l|}{\textbf{\#Testing Samples}} & \multicolumn{1}{l|}{\textbf{\#Category}} \\ \hline
MNIST     & {}$28\times28$, gray{}  & 60,000                                     & 10,000                                    & 10                              \\ \hline
SVHN      & {}$32\times32$, color{} & 73,000                                     & 26,000                                    & 10                              \\ \hline
CIFAR-10  & {}$32\times32$, color{} & 50,000                                     & 10,000                                    & 10                              \\ \hline
N-MNIST  & {}$34\times34\times2$, ON and OFF spikes{} & 60,000                                     & 10,000                                    & 10                              \\ \hline
\end{tabular}
\end{table}

\subsubsection{Network Topologies}
We use various SNN architectures depending on the complexity of the benchmark datasets. For MNIST and N-MNIST datasets, we used a network consisting of two sets of alternating convolutional and spatial-pooling layers followed by two fully-connected layers. This network architecture is derived from LeNet5 model \cite{lecun1998gradient}. Note that table \ref{table:network} summarizes the layer type, kernel size, the number of output feature maps and stride of SNN model for MNIST dataset. The kernel size shown in the table is for 3-D convolution where the 1\textsuperscript{st} dimension is for number of input feature-maps and 2\textsuperscript{nd}-3\textsuperscript{rd} dimensions are for convolutional kernels. For SVHN and CIFAR-10 datasets, we used deeper network models consisting of 7 to 11 trainable layers including convolutional, spatial-pooling and fully-connected layers. In particular, these networks consisting of beyond 5 trainable layers are constructed using small ($3\times3$) convolutional kernels. We term the deep convolutional SNN architecture that includes $3\times3$ convolutional kernel \cite{simonyan2014very} without residual connections as `VGG SNN' and with skip (residual) connections \cite{he2016deep} as `Residual SNN'. In Residual SNNs, some convolutional layers convolve kernel with the stride of 2 in both \textit{x} and \textit{y} directions, to incorporate the functionality of spatial-pooling layers. Please, refer to tables \ref{table:network} and \ref{table:deepnetwork} that summarize the details of deep convolutional SNN architectures. In the results section, we will discuss the benefit of deep SNNs in terms of classification performance as well as inference speedup and energy efficiency.\par

\begin{table}[h]
\centering
\caption{The deep convolutional spiking neural network architectures for MNIST, N-MNIST and SVHN dataset}
\label{table:network}
\resizebox{\textwidth}{!}{
\begin{tabular}{|c|c|c|c|c|c|c|c|c|c|c|c}
\hline

\multicolumn{4}{|c|}{\textbf{4 layer network}}                          & \multicolumn{4}{c|}{\textbf{VGG7}}                            & \multicolumn{4}{c|}{\textbf{ResNet7}}                                                \\ \hline
\textbf{Layer type}      & \textbf{Kernel size} & \textbf{\#o/p feature-maps} & \textbf{Stride} & \textbf{Layer type}            & \textbf{Kernel size} & \textbf{\#o/p feature-maps} & \textbf{Stride} & \textbf{Layer type}         & \textbf{Kernel size} & \textbf{\#o/p feature-maps} & \multicolumn{1}{l|}{\textbf{Stride}} \\ \hline
Convolution     & 1$\times$5$\times$5 & 20   & 1 & Convolution         & 3$\times$3$\times$3  & 64   & 1 & Convolution         & 3$\times$3$\times$3  & 64   & \multicolumn{1}{l|}{1} \\
Average-pooling & 2$\times$2 &   & 2 & Convolution     & 64$\times$3$\times$3    & 64 & 2 & Average-pooling             & 2$\times$2    &      & \multicolumn{1}{l|}{2} \\
         &    &      &  & Average-pooling   &  2$\times$2 &    & 2 &                     &        &      & \multicolumn{1}{l|}{}  \\ \hline

Convolution     & 20$\times$5$\times$5 & 50  & 1 & Convolution         & 64$\times$3$\times$3  & 128  & 1 & Convolution         & 64$\times$3$\times$3  & 128  & \multicolumn{1}{l|}{1} \\
Average-pooling & 2$\times$2           &     & 2 & Convolution  & 128$\times$3$\times$3  & 128  & 2 & Convolution         & 128$\times$3$\times$3  & 128  & \multicolumn{1}{l|}{2} \\
     &  &  &  &  Convolution & 128$\times$3$\times$3  & 128  & 2 & Skip convolution & 64$\times$1$\times$1 & 128  & \multicolumn{1}{l|}{2} \\
         &      &     &   & Average-pooling   &   2$\times$2    &    & 2 &                     &        &      & \multicolumn{1}{l|}{}  \\ \hline

                              & & &  &          &    &    &  & Convolution         & 128$\times$3$\times$3  & 256  & \multicolumn{1}{l|}{1} \\
                              & & &  &          &    &    &  & Convolution         & 256$\times$3$\times$3  & 256  & \multicolumn{1}{l|}{2} \\
   							  & & &  &          &    &    &  & Skip convolution& 128$\times$1$\times$1  & 256 & \multicolumn{1}{l|}{2} \\  \hline

Fully-connected &       & 200 &   & Fully-connected     &        & 1024 &   & Fully-connected     &        & 1024 & \multicolumn{1}{l|}{}  \\
Output  &       & 10   &   & Output  &        & 10   &   & Output  &        & 10   & \multicolumn{1}{l|}{}  \\ \hline
\end{tabular}}
\end{table}

\begin{table}[h]
\centering
\caption{The deep convolutional spiking neural network architectures for a CIFAR-10 dataset}
\label{table:deepnetwork}
\resizebox{\textwidth}{!}{
\begin{tabular}{|c|c|c|c|c|c|c|c|c|c|c|c}
\hline
\multicolumn{4}{|c|}{\textbf{VGG9}}                          & \multicolumn{4}{c|}{\textbf{ResNet9}}                            & \multicolumn{4}{c|}{\textbf{ResNet11}}                                                \\ \hline
\textbf{Layer type}      & \textbf{Kernel size} & \textbf{\#o/p feature-maps} & \textbf{Stride} & \textbf{Layer type}            & \textbf{Kernel size} & \textbf{\#o/p feature-maps} & \textbf{Stride} & \textbf{Layer type}         & \textbf{Kernel size} & \textbf{\#o/p feature-maps} & \multicolumn{1}{l|}{\textbf{Stride}} \\ \hline
Convolution     & 3$\times$3$\times$3 & 64   & 1 & Convolution         & 3$\times$3$\times$3  & 64   & 1 & Convolution         & 3$\times$3$\times$3  & 64   & \multicolumn{1}{l|}{1} \\
Convolution     & 64$\times$3$\times$3 & 64  & 1 & Average-pooling     & 2$\times$2    &      & 2 & Average-pooling          & 2$\times$2    &      & \multicolumn{1}{l|}{2} \\
Average-pooling         & 2$\times$2   &      & 2 &                     &        &      &   &                     &        &      & \multicolumn{1}{l|}{}  \\ \hline

Convolution     & 64$\times$3$\times$3 & 128  & 1 & Convolution         & 64$\times$3$\times$3   & 128  & 1 & Convolution         & 64$\times$3$\times$3  & 128  & \multicolumn{1}{l|}{1} \\
Convolution     & 128$\times$3$\times$3& 128  & 1 & Convolution         & 128$\times$3$\times$3  & 128  & 1 & Convolution         & 128$\times$3$\times$3 & 128  & \multicolumn{1}{l|}{1} \\
Average-pooling         & 2$\times$2    &    &  2 & Skip convolution & 64$\times$1$\times$1   & 128  & 1 & Skip convolution & 64$\times$1$\times$1  & 128  & \multicolumn{1}{l|}{1} \\ \hline

Convolution     & 128$\times$3$\times$3 & 256  & 1 & Convolution         & 128$\times$3$\times$3  & 256  & 1 & Convolution         & 128$\times$3$\times$3  & 256  & \multicolumn{1}{l|}{1} \\
Convolution     & 256$\times$3$\times$3 & 256  & 1 & Convolution         & 256$\times$3$\times$3  & 256  & 2 & Convolution         & 256$\times$3$\times$3  & 256  & \multicolumn{1}{l|}{2} \\
Convolution     & 256$\times$3$\times$3 & 256  & 1 & Skip connection & 128$\times$1$\times$1  & 256  & 2 & Skip convolution & 128$\times$1$\times$1 & 256  & \multicolumn{1}{l|}{2} \\
Average-pooling         & 2$\times$2     &     & 2 &                     &          &      &   &                     &        &      & \multicolumn{1}{l|}{}  \\ \hline

                              & & &  & Convolution         & 256$\times$3$\times$3  & 512  & 1 & Convolution         & 256$\times$3$\times$3  & 512  & \multicolumn{1}{l|}{1} \\
   							  & & &  & Convolution         & 512$\times$3$\times$3  & 512  & 2 & Convolution         & 512$\times$3$\times$3  & 512  & \multicolumn{1}{l|}{1} \\
   							  & & &  & Skip convolution & 256$\times$1$\times$1  & 512  & 2 & Skip convolution & 512$\times$1$\times$1 & 512  & \multicolumn{1}{l|}{1} \\ \hline

                &       &      &   &                     &        &      &       & Convolution         & 512$\times$3$\times$3  & 512  & \multicolumn{1}{l|}{1} \\
                &       &      &   &                     &        &      &       & Convolution         & 512$\times$3$\times$3  & 512  & \multicolumn{1}{l|}{2} \\
                &       &      &   &                     &        &      &       & Skip convolution & 512$\times$1$\times$1 & 512  & \multicolumn{1}{l|}{2} \\ \hline

Fully-connected &       & 1024 &   & Fully-connected     &        & 1024 &   & Fully-connected     &        & 1024 & \multicolumn{1}{l|}{}  \\
Output &       & 10   &   & Output  &        & 10   &   & Output  &        & 10   & \multicolumn{1}{l|}{}  \\ \hline
\end{tabular}}
\end{table}

\subsubsection{ANN-SNN Conversion Scheme}
\label{ANNSNN}
As mentioned previously, off-the-shelf trained ANNs can be successfully converted to SNNs by replacing ANN (ReLU) neurons with Integrate and Fire (IF) spiking neurons and adjusting the neuronal thresholds with respect to synaptic weights. 
In the literature, several methods have been proposed \cite{cao2015spiking,diehl2016conversion,hunsberger2015spiking,sengupta2018going,rueckauer2017conversion} for balancing appropriate ratios between neuronal thresholds and synaptic weights of spiking neuron in the case of ANN-SNN conversion. In this paper, we compare various aspects of our direct-spike trained models with two prior ANN-SNN conversion works \cite{sengupta2018going, diehl2016conversion}, which proposed near-lossless ANN-SNN conversion schemes for deep network architectures. The first scheme \cite{sengupta2018going} balanced the neuronal firing thresholds with respect to corresponding synaptic weights layer-by-layer depending on the actual spiking activities of each layer using a subset of training samples. The second scheme \cite{diehl2016conversion} balanced the neuronal firing thresholds with the consideration of ReLU activations in the corresponding ANN layer. Basically, we compare our direct-spike trained model with converted SNNs on the same network architecture in terms of accuracy, inference speed and energy-efficiency. Please note that there are a couple of differences on the network architecture between the conversion networks \cite{sengupta2018going,diehl2016conversion} and our scheme. First, the conversion networks always use average-pooling to reduce the size of previous convolutional output feature-map, whereas our models interchangeably use average pooling or convolve kernels with a stride of 2 in the convolutional layer. Next, the conversion networks only consider identity skip connections for residual SNNs. However, we implement skip connections using either identity mapping or $1\times1$ convolutional kernel.

\subsubsection{Spike Generation Scheme}
\label{Spike Gen}
For the static vision datasets (MNIST, SVHN and CIFAR-10), each input pixel intensity is converted to a stream of Poisson-distributed spike events that have equivalent firing rates. Specifically, at each time step, the pixel intensity is compared with a uniformly distributed random number (in the range between 0 and 1). If pixel intensity is greater than the random number at the corresponding time step, a spike is generated. This rate-based spike encoding is used to feed the input spikes to the network for a given period of time during both training and inference. For color image datasets, we use the pre-processing technique of horizontal flip before generating input spikes. These input pixels are normalized to represent zero mean and unit standard deviation. Thereafter, we scale the pixel intensities to bound them in the range [-1,1] to represent the whole spectrum of input pixel representations. The normalized pixel intensities are converted to Poisson-distributed spike events such that the generated input signals are bipolar spikes. For the neuromorphic version of the dataset (N-MNIST), we use the original (unfiltered and uncentered) version of spike streams to directly train and test the network in the time domain.

\subsubsection{Time-steps}
As mentioned in section \ref{Spike Gen}, we generate a stochastic Poisson-distributed spike train for each input pixel intensity for event-based operation. The duration of the spike train is very important for SNNs. We measure the length of the spike train (spike time window) in time-steps. For example, a 100 time-step spike train will have approximately 50 random spikes if the corresponding pixel intensity is half in a range of [0,1]. If the number of time-steps (spike time window) is too less, then the SNN will not receive enough information for training or inference. On the other hand, a large number of time-steps will destroy the stochastic property of SNNs and get rid of noise and imprecision at the cost of high latency and power consumption. Hence, the network will not have much energy efficiency over ANN implementations. For these reasons, we experimented with the different number of time-steps to empirically obtain the optimal number of time-steps required for both training and inference. The experimental process and results are explained in the following subsections.

\begin{figure*}[h]
\centering
\subfloat[]{\label{fig4a}
  \centering
  \includegraphics[width=.5\textwidth]{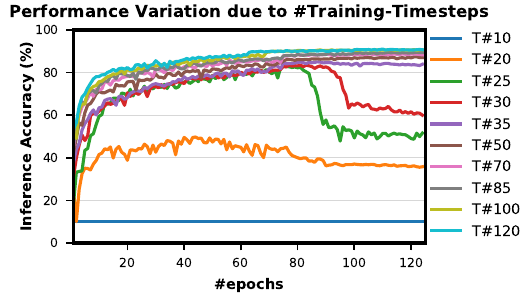}}
\subfloat[]{\label{fig4b}
  \centering
  \includegraphics[width=.5\textwidth]{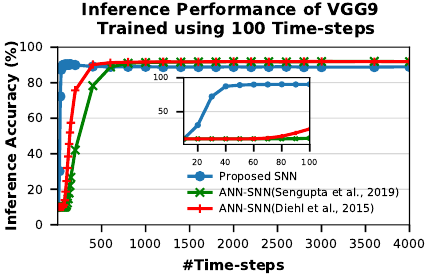}}
\caption{Inference performance variation due to (a) \#Training-Timesteps and (b) \#Inference-Timesteps. \textit{T\#} in (a) indicates number of time-steps used for training.
Figure (a) shows that inference accuracy starts to saturate as \#training-timesteps increase. In (b), the zoomed version on the inset shows that the SNN trained with the proposed scheme performs very well even with only 30 time-steps while the peak performance occurs around 100 time-steps.
}
\label{fig4}
\end{figure*}

\subsubsection*{Optimal \#time-steps for Training}
\label{timestepstraining}
A spike event can only represent 0 or 1 in each time step, therefore usually its bit precision is considered 1. However, the spike train provides temporal data, which is an additional source of information. Therefore, the spike train length (number of time-steps) in SNN can be considered as its actual precision of neuronal activation. To obtain the optimal \#time-steps required for our proposed training method, we trained VGG9 networks on CIFAR-10 dataset using different time-steps ranging from 10 to 120 (shown in figure \ref{fig4a}). We found that for only 10 time-steps, the network is unable to learn anything as there is not enough information (input precision too low) for the network to be able to learn. This phenomenon is explained by the lack of spikes in the final output. With the initial weights, the accumulated sum of the LIF neuron is not enough to generate output spikes in the later layers. Hence, none of the input spikes propagates to the final output neurons and the output distributions remain 0. Therefore, the computed gradients are always 0 and the network is not updated. For 35-50 time-steps, the network learns well and converges to a reasonable point. From 70 time-steps, the network accuracy starts to saturate. At about 100 time-steps, the network training improvement completely saturates. This is consistent with the bit precision of the inputs. It has been shown in \cite{sarwar2018energy} that 8 bit inputs and activations are sufficient to achieve optimal network performance for standard image recognition tasks.
Ideally, we need 128 time-steps to represent 8 bit inputs using bipolar spikes.  However, 100 time-steps proved to be sufficient as more time-steps provide marginal improvement. We observe a similar trend in VGG7, ResNet7, ResNet9 and ResNet11 SNNs as well while training for SVHN and CIFAR-10 datasets. Therefore, we considered 100 time-steps as the optimal \#time-steps for training in our proposed methodology. Moreover, for MNIST dataset, we used 50 time-steps since the required bit precision is only 4 bits \cite{sarwar2018energy}.

\subsubsection*{Optimal \#time-steps for Inference}
To obtain the optimal \#time-steps required for inferring an image utilizing a network trained with our proposed method, we conducted similar experiments as described in section \mbox{\ref{timestepstraining}}. We first trained a VGG9 network for CIFAR-10 dataset using 100 time-steps (optimal according to experiments in section \mbox{\ref{timestepstraining}}). Then, we tested the network performances with different time-steps ranging from 10 to 4000 (shown in figure \ref{fig4b}). We observed that the network performs very well even with only 30 time-steps while the peak performance occurs around 100 time-steps. For more than 100 time-steps, the accuracy degrades slightly from the peak. This behavior is very different from ANN-SNN converted networks where the accuracy keeps on improving as \#time-steps is increased (shown in figure \ref{fig4b}). This can be attributed to the fact that our proposed spike-based training method incorporates the temporal information well into the network training procedure so that the trained network is tailored to perform best at a specific spike time window for the inference. On the other hand, the ANN-SNN conversion schemes are unable to incorporate the temporal information of the input in the trained network. Hence, the ANN-SNN conversion schemes require much higher \#time-steps (compared to SNN trained using the proposed method) for the inference in order to resemble input-output mappings similar to ANNs.

\subsection{Results}
In this section, we analyze the classification performance and efficiency achieved by the proposed spike-based training methodology for deep convolutional SNNs compared to the performance of the transformed SNN using ANN-SNN conversion scheme.

\subsubsection{The Classification Performance}
Most of the classification performances available in the literature for SNNs are for MNIST and CIFAR-10 datasets. The popular methods for SNN training are `Spike Time Dependent Plasticity (STDP)' based unsupervised learning \cite{diehl2015unsupervised,brader2007learning,srinivasan2018spilinc,srinivasan2018stdp} and `spike-based backpropagation' based supervised learning \cite{lee2016training,jin2018hybrid,wu2018spatio,neftci2017event,mostafa2017supervised}. There are a few works \cite{tavanaei2016bio,kheradpisheh2016stdp,tavanaei2017multi,lee2018training} which tried to combine the two approaches to get the best of both worlds. However, these training methods were able to neither train deep SNNs nor achieve good inference performance compared to ANN implementations. Hence, ANN-SNN conversion schemes have been explored by researchers \cite{cao2015spiking,diehl2016conversion,hunsberger2015spiking,sengupta2018going,rueckauer2017conversion}. Till date, ANN-SNN conversion schemes achieved the best inference performance for CIFAR-10 dataset using deep networks \cite{sengupta2018going,rueckauer2017conversion}. Classification performances of all these works are listed in table \ref{table:acc} along with ours. To the best of our knowledge, we achieved the best inference accuracy for MNIST using LeNet structured network compared to our spike based training approaches. We also achieved accuracy performance comparable with ANN-SNN converted network \cite{sengupta2018going,diehl2015fast} for CIFAR-10 dataset while beating all other spike-based training methods.

\begin{table}[h]
\centering
\caption{Comparison of the SNNs classification accuracies on MNIST, N-MNIST and CIFAR-10 datasets.}
\label{table:acc}
\resizebox{\textwidth}{!}{
\begin{tabular}{llllll}
\cline{1-5}
\multicolumn{1}{|l|}{\textbf{Model}}        & \multicolumn{1}{l|}{\textbf{Learning Method}}     & \multicolumn{1}{l|}{\textbf{\begin{tabular}[c]{@{}l@{}}Accuracy\\ (MNIST)\end{tabular}}} &  \multicolumn{1}{l|}{\textbf{\begin{tabular}[c]{@{}l@{}}Accuracy\\ (N-MNIST)\end{tabular}}} & \multicolumn{1}{l|}{\textbf{\begin{tabular}[c]{@{}l@{}}Accuracy\\ (CIFAR-10)\end{tabular}}} &  \\ \cline{1-5}

\multicolumn{1}{|l|}{Hunsberger et al.\cite{hunsberger2015spiking}}                  & \multicolumn{1}{l|}{Offline learning, conversion} & \multicolumn{1}{l|}{98.37\%} & \multicolumn{1}{l|}{--}& \multicolumn{1}{l|}{82.95\%}  &  \\ \cline{1-5}

\multicolumn{1}{|l|}{Esser et al.\cite{esser2016convolutional}}        & \multicolumn{1}{l|}{Offline learning, conversion} & \multicolumn{1}{l|}{--} & \multicolumn{1}{l|}{--} & \multicolumn{1}{l|}{89.32\%}            &  \\ \cline{1-5}

\multicolumn{1}{|l|}{Diehl et al.\cite{diehl2016conversion}}                     & \multicolumn{1}{l|}{Offline learning, conversion}  & \multicolumn{1}{l|}{99.10\%} & \multicolumn{1}{l|}{--} & \multicolumn{1}{l|}{--}          &  \\ \cline{1-5}

\multicolumn{1}{|l|}{Rueckauer et al.\cite{rueckauer2017conversion}}                     & \multicolumn{1}{l|}{Offline learning, conversion}  & \multicolumn{1}{l|}{99.44\%} & \multicolumn{1}{l|}{--} & \multicolumn{1}{l|}{88.82\%}           &  \\ \cline{1-5}

\multicolumn{1}{|l|}{Sengupta et al.\cite{sengupta2018going}}                     & \multicolumn{1}{l|}{Offline learning, conversion}  & \multicolumn{1}{l|}{--} & \multicolumn{1}{l|}{--} & \multicolumn{1}{l|}{91.55\%}           &  \\ \cline{1-5}

\multicolumn{1}{|l|}{Kheradpisheh et al.\cite{kheradpisheh2016stdp}}              & \multicolumn{1}{l|}{Layerwise STDP + offline SVM classifier}        & \multicolumn{1}{l|}{98.40\%}  & \multicolumn{1}{l|}{--} & \multicolumn{1}{l|}{--}         &  \\ \cline{1-5}

\multicolumn{1}{|l|}{Panda et al.\cite{panda2016unsupervised}}                & \multicolumn{1}{l|}{Spike-based autoencoder}    & \multicolumn{1}{l|}{99.08\%}   & \multicolumn{1}{l|}{--} & \multicolumn{1}{l|}{70.16\%}       &  \\ \cline{1-5}

\multicolumn{1}{|l|}{Lee et al.\cite{lee2016training}}                     & \multicolumn{1}{l|}{Spike-based BP}              & \multicolumn{1}{l|}{99.31\%}  & \multicolumn{1}{l|}{98.74\%}& \multicolumn{1}{l|}{--}         &  \\ \cline{1-5}

\multicolumn{1}{|l|}{Wu et al.\cite{wu2018spatio}}                     & \multicolumn{1}{l|}{Spike-based BP}              & \multicolumn{1}{l|}{99.42\%}   & \multicolumn{1}{l|}{98.78\%}  & \multicolumn{1}{l|}{50.70\%}      &  \\ \cline{1-5}

\multicolumn{1}{|l|}{Lee et al.\cite{lee2018training}}                & \multicolumn{1}{l|}{STDP-based pretraining + spike-based BP} & \multicolumn{1}{l|}{99.28\%}  & \multicolumn{1}{l|}{--}  & \multicolumn{1}{l|}{--}       &  \\ \cline{1-5}

\multicolumn{1}{|l|}{Jin et al.\cite{jin2018hybrid}}                   & \multicolumn{1}{l|}{Spike-based BP}              & \multicolumn{1}{l|}{99.49\%} & \multicolumn{1}{l|}{98.88\%}& \multicolumn{1}{l|}{--}          &  \\ \cline{1-5}

\multicolumn{1}{|l|}{Wu et al.\cite{wu2018direct}}                   & \multicolumn{1}{l|}{Spike-based BP}              & \multicolumn{1}{l|}{--} & \multicolumn{1}{l|}{99.53\%} & \multicolumn{1}{l|}{90.53\%}          &  \\ \cline{1-5}

\multicolumn{1}{|l|}{This work}                 & \multicolumn{1}{l|}{Spike-based BP}              & \multicolumn{1}{l|}{99.59\%}  & \multicolumn{1}{l|}{99.09\%} & \multicolumn{1}{l|}{90.95\%}        &  \\ \cline{1-5}

                                          &                                              &                                                   &                                        & 
\end{tabular}}
\end{table}

For a more extensive comparison, we compare the inference performances of trained networks using our proposed methodology with the ANNs and ANN-SNN conversion scheme for same network configuration (depth and structure) side by side in table \ref{table:AccuracyPerfm}. We also compare with the previous best SNN training results found in the literature that may or may not have the same network depth and structure as ours. The ANN-SNN conversion scheme is a modified and improved version of \cite{sengupta2018going}. We are using this modified scheme since it achieves better conversion performance than \cite{sengupta2018going} as explained in section \ref{ANNSNN}. Note that all reported classification accuracies are the average of the maximum inference accuracies for 3 independent runs with different seeds.\par

After initializing the weights, we train the SNNs using a spike-based BP algorithm in an end-to-end manner with Poisson-distributed spike train inputs. Our evaluation of MNIST dataset yields a classification accuracy of 99.59\%, which is the best compared to any other SNN training scheme and also identical to other ANN-SNN conversion schemes. We achieve \texttildelow96\% inference accuracy on SVHN dataset for both trained non-residual and residual SNN. 
Inference performance for SNNs trained on SVHN dataset has not been reported previously in the literature. We implemented three different networks, as shown in table \ref{table:deepnetwork}, for classifying CIFAR-10 dataset using a proposed spike-based BP algorithm. For the VGG9 network, the ANN-SNN conversion schemes provides a near lossless converted network compared to baseline ANN implementation while our proposed training method yields a classification accuracy of 90.45\%. 
For ResNet9 network, the ANN-SNN conversion schemes provide inference accuracy within 0.5-1\% of baseline ANN implementation. However, our proposed spike-based training method achieves inference accuracy that is within \texttildelow1.5\% of baseline ANN implementation. In the case of ResNet11, we observe that the inference accuracy improvement is marginal compared to ResNet9 for baseline ANN implementation and ANN-SNN conversion schemes. However, our proposed SNN training shows improvement of \texttildelow0.5\% for ResNet11 compared to ResNet9. Overall, our proposed training method achieves comparable inference accuracies for both ResNet and VGG networks compared to baseline ANN implementation and ANN-SNN conversion schemes.

\begin{table}[h]
\centering
\caption{Comparison of classification performance}
\label{table:AccuracyPerfm}
\begin{tabular}{|c|c|c|c|c|c|c|} 
\hline
\multicolumn{7}{|c|}{\textbf{Inference Accuracy}}                                                                                                                                      \\ 
\hline
\multirow{2}{*}{\textbf{Dataset}}  & \multirow{2}{*}{\textbf{Model}} & \multirow{2}{*}{\textbf{ANN}} & \textbf{ANN-SNN}~             & \textbf{ANN-SNN}~                & \textbf{SNN}~                     & \textbf{SNN}~        \\
                          &                        &                      & \textbf{(Diehl et al. }\cite{diehl2015fast}\textbf{)} & \textbf{(Sengupta et al. }\cite{sengupta2018going}\textbf{)}  & \textbf{(Previous Best)}          & \textbf{(This Work)}  \\ 
\hline
MNIST                     & LeNet                  & 99.57\%              & 99.55\%              & 99.59\%                 & 99.49\%\cite{jin2018hybrid}                  & 99.59\%     \\ 
\hline
N-MNIST                   & LeNet                  & –                    & –                    & –                       & 99.53\%\cite{wu2018direct}                  & 99.09\%     \\ 
\hline
\multirow{2}{*}{SVHN}     & VGG7                   & 96.36\%              & 96.33\%              & 96.30\%                 & –                        & 96.06\%     \\ 
\cline{2-7}
                          & ResNet7                & 96.43\%              & 96.33\%              & 96.40\%                 & –                        & 96.21\%     \\ 
\hline
\multirow{3}{*}{CIFAR-10} & VGG9                   & 91.98\%              & 91.89\%              & 92.01\%                 & \multirow{3}{*}{90.53\%\cite{wu2018direct}} & 90.45\%     \\ 
\cline{2-5}\cline{7-7}
                          & ResNet9                & 91.85\%              & 90.78\%              & 91.59\%                 &                          & 90.35\%     \\ 
\cline{2-5}\cline{7-7}
                          & ResNet11               & 91.87\%              & 90.98\%              & 91.65\%                 &                          & 90.95\%     \\
\hline
\end{tabular}
\end{table}

\subsubsection{Accuracy Improvement with Network Depth}
\begin{figure*}[h]
\centering
\subfloat[]{\label{fig5a}
  \centering
  \includegraphics[width=.5\textwidth]{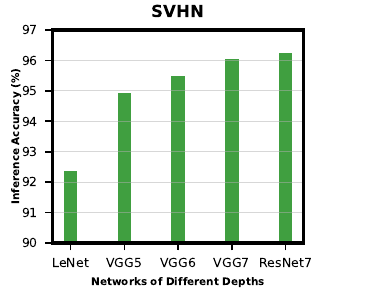}}
\subfloat[]{\label{fig5b}
  \centering
  \includegraphics[width=.5\textwidth]{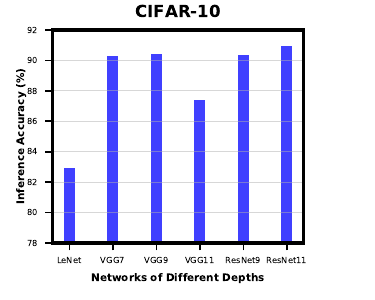}}
\caption{Accuracy improvement with network depth for (a) SVHN dataset and (b) CIFAR-10 dataset. In (a), inference accuracy improves with an increase in network depth. In (b), the non-residual networks saturate at a certain depth and start to degrade if network depth increases further. However, the residual blocks in deep residual ANNs allow the network to maintain peak classification accuracy (ResNet9 and ResNet11).}
\label{fig5}
\end{figure*}

In order to analyze the effect of network depth for SNNs, we experimented with networks of different depths while training for SVHN and CIFAR-10 datasets. For SVHN dataset, we started with a shallow network derived from LeNet5 model \cite{lecun1998gradient} with 2 convolutional and 2 fully-connected layers. This network was able to achieve inference accuracy of only 92.38\%. Then, we increased the network depth by adding 1 convolutional layer before the 2 fully-connected layers and we termed this network as VGG5. VGG5 network was able to achieve significant improvement over its predecessor. Similarly, we tried VGG6 followed by VGG7, and the improvement started to become very small. We have also trained ResNet7 to understand how residual networks perform compared to non-residual networks of similar depth. The results of these experiments are shown in figure \ref{fig5a}. We carried out similar experiments for CIFAR-10 dataset as well. The results show a similar trend as described in figure \ref{fig5b}. These results ensure that network depth improves the learning capacity of direct-spike trained SNNs similar to ANNs. The non-residual networks saturate at a certain depth and start to degrade if network depth is further increased (VGG11 in figure \ref{fig5b}) due to the degradation problem mentioned in \cite{he2016deep}. In such a scenario, the residual connections in deep residual ANNs allow the network to maintain peak classification accuracy utilizing the skip connections \cite{he2016deep}, as seen in figure \ref{fig5b} (ResNet9 and ResNet11).

\setcounter{subfigure}{0}

\section{Discussion}
\subsection{Comparison with Relevant works}
In this section, we compare our proposed supervised learning algorithm with other recent spike-based BP algorithms. The spike-based learning rules primarily focus on directly training and testing SNNs with spike-trains, and no conversion is necessary for applying in real-world spiking scenario. In recent years, there are an increasing number of supervised gradient descent method in spike-based learning. The \cite{panda2016unsupervised} developed a spike-based auto-encoder mechanism to train deep convolutional SNNs. They dealt with membrane potential as a differentiable signal and showed recognition capabilities in standard vision tasks (MNIST and CIFAR-10 datasets). Meanwhile, \cite{lee2016training} followed the approach using differentiable membrane potential to explore a spike-based BP algorithm in an end-to-end manner. In addition, \cite{lee2016training} presented the error normalization scheme to prevent exploding gradient phenomena while training deep SNNs. Researchers in \cite{jin2018hybrid} proposed hybrid macro/micro level backpropagation (HM2-BP). HM2-BP is developed to capture the temporal effect of the individual spike (in micro-level) and rate-encoded error (at macro-level). The reference \cite{shrestha2018slayer} employed exponential function for the approximate derivative of neuronal function and developed a credit assignment scheme to calculate the temporal dependencies of error throughout the layers. \cite{huh2018gradient} has trained recurrent spiking networks by replacing the threshold with a gate function and employing BPTT technique \cite{werbos1990backpropagation}. While BPTT technique has been a popular method to train recurrent artificial and spiking recurrent networks, \cite{lillicrap2019backpropagation} points out the storing and retrieving past variables and differentiation them through time in biological neurons seems to be impossible. Recently, e-prop \cite{bellec2019solution} presented an approximation method to bypass neuronal state savings for enhancing the computational efficiency of BPTT. In temporal spike encoding domain, \cite{mostafa2017supervised} proposed an interesting temporal spike-based BP algorithm by treating the spike-time as the differential activation of neuron. Temporal encoding based SNN has the potential to process the tasks with the small number of spikes. All of these works demonstrated spike-based learning in simple network architectures and has a large gap in classification accuracy compared to deep ANNs. More recently, \cite{wu2018direct} presented a neuron normalization technique (called NeuNorm) that calculates the average input firing rates to adjust neuron selectivity. NeuNorm enables spike-based training within a relatively short time-window while achieving competitive performances. In addition, they presented an input encoding scheme that receives both spike and non-spike signals for preserving the precision of input data. 

There are several points that distinguish our work from others. First, we use a pseudo derivative method that accounts for leaky effect in membrane potential of LIF neurons. We approximately estimate the leaky effect by comparing total membrane potential value and obtain the ratio between IF and LIF neurons. During the back-propagating phase, the pseudo derivative of LIF neuronal function is estimated by combining the straight through estimation and leak correctional term as described in equation (22). Next, we construct our networks by leveraging frequently used architectures such as VGG \cite{simonyan2014very} and ResNet \cite{he2016deep}. To the best of our knowledge, this is the first work that demonstrates spike-based supervised BP learning for SNNs containing more than 10 trainable layers. Our deep SNNs obtain the superior classification accuracies in MNIST, SVHN and CIFAR-10 datasets in comparison to the other networks trained with the spike-based algorithm. In addition, as opposed to complex error or neuron normalization method adopted by \cite{lee2016training} and \cite{wu2018direct}, respectively, we demonstrate that deep SNNs can be naturally trained by only accounting for spiking activities of the network. As a result, our work paves the effective way for training deep SNNs with a spike-based BP algorithm.

\subsection{Spike Activity Analysis}
The most important advantage of event-based operation of neural networks is that the events are very sparse in nature. To verify this claim, we analyzed the spiking activities of the direct-spike trained SNNs and ANN-SNN converted networks in the following subsections.

\subsubsection{Spike Activity per Layer}
\label{SA_layer}
\begin{figure*}[h]
\centering
\subfloat[]{\label{fig6a}
  \centering
  \includegraphics[width=.48\textwidth]{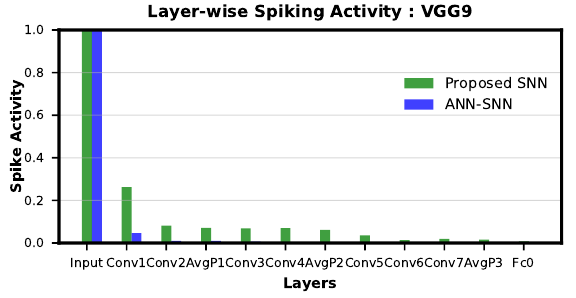}}
\subfloat[]{\label{fig6b}
  \centering
  \includegraphics[width=.48\textwidth]{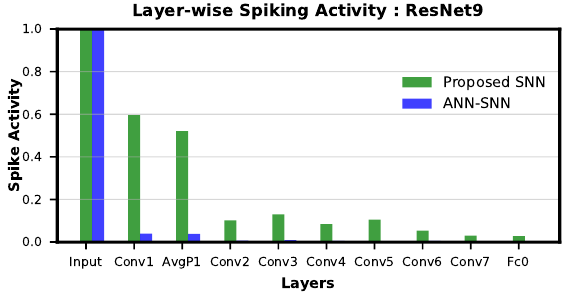}}
\caption{Layer-wise spike activity in direct-spike trained SNN and ANN-SNN converted network for CIFAR-10 dataset: (a) VGG9 (b) ResNet9 network. The spike activity is normalized with respect to the input layer spike activity, which is the same for both networks. The spike activity reduces significantly for both SNN and ANN-SNN converted network towards the later layers. We have used scheme 1 for ANN-SNN conversion.
}\label{fig6}
\end{figure*}

The layer-wise spike activities of both SNN trained using our proposed methodology, and ANN-SNN converted network (using scheme 1) for VGG9 and ResNet9 are shown in figure \ref{fig6a} and \ref{fig6b}, respectively. In the case of ResNet9, only the first average pooling layer's output spike activity is shown in the figure as for the direct-spike trained SNN, the other spatial-poolings are done by stride 2 convolutions. In figure \ref{fig6}, it can be seen that the input layer has the highest spike activity that is significantly higher than any other layer. The spike activity reduces significantly as the network depth increases.

We can observe from figure \ref{fig6a} and figure \ref{fig6b} that the average spike activity in a direct-spike trained SNN is much higher than ANN-SNN converted network. The ANN-SNN converted network uses a higher threshold compared to 1 (in case of direct-spike trained SNN) since the conversion scheme applies layer-wise neuronal threshold modulation. This higher threshold reduces spike activity in ANN-SNN converted networks. However, in both cases, the spike activity decreases with increasing network depth.\par

\setcounter{subfigure}{0}

\begin{table}[h]
\centering
\caption{\#Spikes/Image inference and spike efficiency comparison between SNN and ANN-SNN converted networks for benchmark datasets trained on different network models. (For each network, the 1\textsuperscript{st} row corresponds to iso-accuracy and the 2\textsuperscript{nd} row corresponds to maximum-accuracy condition.)}
\label{Spikes/Inference}
\begin{tabular}{|c|c|c|c|c|c|c|} 
\hline
\multirow{2}{*}{\textbf{Dataset}}  & \multirow{2}{*}{\textbf{Model}}    & \multicolumn{3}{c|}{\textbf{Spike/Image}}                                              & \multicolumn{2}{c|}{\textbf{Spike Efficiency Compared to}}  \\ 
\cline{3-7}
                          &                           & \textbf{SNN}    & \textbf{ANN-SNN} \cite{diehl2015fast}                 & \textbf{ANN-SNN} \cite{sengupta2018going}                & \textbf{ANN-SNN} \cite{diehl2015fast}               & \textbf{ANN-SNN} \cite{sengupta2018going}                  \\ 
\cline{4-7}
\hline
\multirow{2}{*}{MNIST}    & \multirow{2}{*}{LeNet}    & \multirow{2}{*}{5.52E+04}   & \multirow{2}{*}{3.4E+04} & 2.9E+04                 & \multirow{2}{*}{0.62x} & 0.53x                     \\ 
\cline{5-5}\cline{7-7}
                          &                           &                          &                          & 7.3E+04                 &                        & 1.32x                     \\ 
\hline
\multirow{4}{*}{SVHN}     & \multirow{2}{*}{VGG7}     & \multirow{2}{*}{5.56E+06} & 3.7E+06                  & 1.0E+07                 & 0.67x                  & 1.84x                     \\ 
\cline{4-7}
                          &                           &                          & 1.9E+07                  & 1.7E+07                 & 3.40x                  & 2.99x                     \\ 
\cline{2-7}
                          & \multirow{2}{*}{ResNet7}  & \multirow{2}{*}{4.66E+06} & 3.9E+06                  & 3.1E+06                 & 0.85x                  & 0.67x                     \\ 
\cline{4-7}
                          &                           &                          & 2.4E+07                  & 2.0E+07                 & 5.19x                  & 4.30x                     \\ 
\hline
\multirow{6}{*}{CIFAR-10} & \multirow{2}{*}{VGG9}     & \multirow{2}{*}{1.24E+06} & 1.6E+06                  & 2.2E+06                 & 1.32x                  & 1.80x                     \\ 
\cline{4-7}
                          &                           &                          & 8.3E+06                  & 9.6E+06                 & 6.68x                  & 7.78x                     \\ 
\cline{2-7}
                          & \multirow{2}{*}{ResNet9}  & \multirow{2}{*}{4.32E+06} & 2.7E+06                  & 1.5E+06                 & 0.63x                  & 0.35x                     \\ 
\cline{4-7}
                          &                           &                          & 1.0E+07                  & 7.8E+06                 & 2.39x                  & 1.80x                     \\ 
\cline{2-7}
                          & \multirow{2}{*}{ResNet11} & \multirow{2}{*}{1.53E+06} & \multirow{2}{*}{9.7E+06} & 1.8E+06                 & \multirow{2}{*}{6.33x} & 1.17x                     \\ 
\cline{5-5}\cline{7-7}
                          &                           &                          &                          & 9.2E+06                 &                        & 5.99x                     \\
\hline
\end{tabular}
\end{table}

\subsubsection{\#Spikes/Inference}
From figure \ref{fig6}, it is evident that average spike activity in ANN-SNN converted networks is much less than in the direct-spike trained SNN. However, for inference, the network has to be evaluated over many time-steps. Therefore, to quantify the actual spike activity for an inference operation, we measured the average number of spikes required for inferring one image. For this purpose, we counted the number of spikes generated (including input spikes) for classifying the test set of a particular dataset for a specific number of time-steps and averaged the count for generating the quantity \textit{`\#spikes per image inference'}. We have used two different time-steps for ANN-SNN converted networks; one for iso-accuracy comparison and the other for maximum accuracy comparison with the direct-spike trained SNNs. Iso-accuracy inference requires less \#time-steps than maximum accuracy inference, hence has a lower number of spikes per image inference. For few networks, the ANN-SNN conversion scheme always provides accuracy less than or equal to the direct-spike trained SNN. Hence, we only compare spikes per image inference in maximum accuracy condition for those ANN-SNN converted networks while comparing with direct-spike trained SNNs. For the analysis, we quantify the spike-efficiency (amount reduction in \#spikes) from the \#spikes/image inference. The results are listed in table \ref{Spikes/Inference}, where the 1\textsuperscript{st} row corresponds to iso-accuracy and the 2\textsuperscript{nd} row corresponds to maximum-accuracy condition for each network. As shown in table \ref{Spikes/Inference}, the direct-spike trained SNNs are more efficient in terms of \#spikes/inference compared to the ANN-SNN converted networks for the maximum accuracy condition. For an iso-accuracy condition, only deep SNNs (such as VGG9 and ResNet11) are more efficient in terms of \#spikes/inference compared to the ANN-SNN converted networks.

\begin{figure*}[h]
\centering
\includegraphics[width=\textwidth]{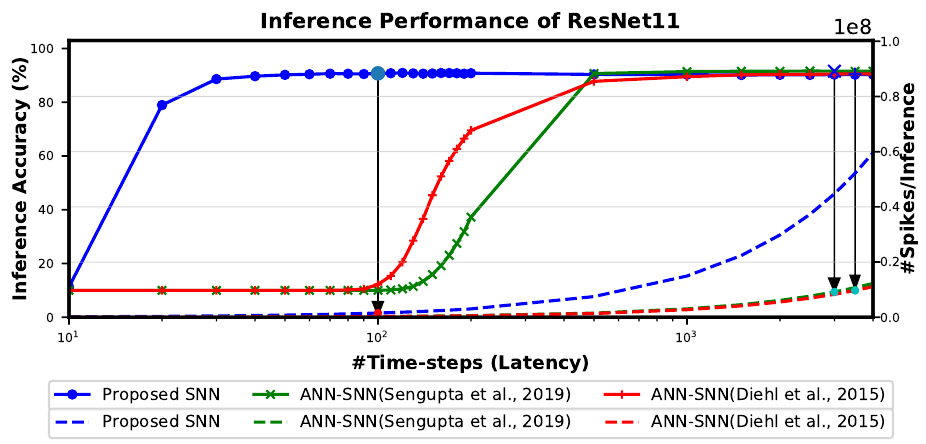}
\caption{The comparison of `accuracy vs latency vs $\#$spikes/inference' for ResNet11 architecture. In this figure, the solid lines are representing inference accuracy while the dashed lines are representing $\#$spikes/inference. The slope of $\#$spikes/inference curve of the proposed SNN is larger than ANN-SNN converted networks. However, since proposed SNN requires much less time-steps for inference, the number of spikes required for one image inference is significantly lower compared to ANN-SNN. The required $\#$time-steps and corresponding $\#$spikes/inference are shown using highlighted points connected by arrows. Log scale is used for x-axis for easier viewing of the accuracy changes for lower number of time-steps.}
\label{fig:fig7}
\end{figure*}

The figure \ref{fig:fig7} shows the relationship between inference accuracy, latency and \#spikes/inference for ResNet11 networks trained on CIFAR-10 dataset. We can observe that \#spikes/inference is higher for direct-spike trained SNN compared to ANN-SNN converted networks at any particular latency. However, SNN trained with spike-based BP requires only 100 time-steps for maximum inference accuracy, whereas ANN-SNN converted networks require 3000-3500 time-steps to reach maximum inference accuracy. Hence, under maximum accuracy condition, direct-spike trained ResNet11 requires much fewer \#spikes/inference compared to ANN-SNN converted networks, while achieving comparable accuracy. Even under iso-accuracy condition, the direct-spike trained ResNet11 requires fewer \#spikes/inference compared to the ANN-SNN converted networks (table \ref{Spikes/Inference}).

\subsection{Inference Speedup}

The time required for inference is linearly proportional to the \#time-steps (figure \ref{fig:fig7}). Hence, we can also quantify the inference speedup for direct-spike trained SNNs compared to ANN-SNN converted networks from the \#time-steps required for inference, as shown in table \ref{InferenceSpeedup}. 
For example, for VGG9 network, the proposed training method can achieve 8x (5x) speedup for iso-accuracy and up to 36x (25x) speedup for maximum accuracy in inference compared to respective ANN-SNN converted networks (i.e., scheme 1 \cite{sengupta2018going} and scheme 2 \cite{diehl2016conversion}). Similarly, for ResNet networks, the proposed training method can achieve 6x speedup for iso-accuracy and up to 35x speedup for maximum accuracy condition in inference. 
It is interesting to note that direct-spike trained SNN is always more efficient in terms of time-steps compared to the equivalent ANN-SNN conversion network, but not in terms of the number of spikes, in some cases. It will require a detailed investigation to determine if ANN-SNN methods used higher firing rates, whether they would be able to classify quickly as well, while incurring a lower number of spike/inference.

\begin{table}[h]
\centering
\caption{Inference \#time-steps and corresponding speedup comparison between SNN and ANN-SNN converted networks for benchmark datasets trained on different network models. (For each network, the 1\textsuperscript{st} row corresponds to iso-accuracy and the 2\textsuperscript{nd} row corresponds to maximum-accuracy condition.)}
\label{InferenceSpeedup}
\begin{tabular}{|c|c|c|c|c|c|c|} 
\hline
\multirow{2}{*}{\textbf{Dataset}}  & \multirow{2}{*}{\textbf{Model}}    & \multicolumn{3}{c|}{\textbf{Timesteps}}                                          & \multicolumn{2}{c|}{\textbf{SNN Inference Speedup Compared to}}  \\ 
\cline{3-7}
                          &                           & \textbf{SNN}    & \textbf{ANN-SNN} \cite{diehl2015fast}                 & \textbf{ANN-SNN} \cite{sengupta2018going}                & \textbf{ANN-SNN} \cite{diehl2015fast}               & \textbf{ANN-SNN} \cite{sengupta2018going}                  \\ 
\cline{4-7}
\hline
\multirow{2}{*}{MNIST}    & \multirow{2}{*}{LeNet}    & \multirow{2}{*}{50}   & \multirow{2}{*}{180}  & 200                     & \multirow{2}{*}{3.6x} & 4x                              \\ 
\cline{5-5}\cline{7-7}
                          &                           &                       &                       & 500                     &                       & 10x                             \\ 
\hline
\multirow{4}{*}{SVHN}     & \multirow{2}{*}{VGG7}     & \multirow{2}{*}{100}  & 500                   & 1600                    & 5x                    & 16x                             \\ 
\cline{4-7}
                          &                           &                       & 2500                  & 2600                    & 25x                   & 26x                             \\ 
\cline{2-7}
                          & \multirow{2}{*}{ResNet7}  & \multirow{2}{*}{100}  & 500                   & 400                     & 5x                    & 4x                              \\ 
\cline{4-7}
                          &                           &                       & 3000                  & 2500                    & 30x                   & 25x                             \\ 
\hline
\multirow{6}{*}{CIFAR-10} & \multirow{2}{*}{VGG9}     & \multirow{2}{*}{100}  & 500                   & 800                     & 5x                    & 8x                              \\ 
\cline{4-7}
                          &                           &                       & 2500                  & 3600                    & 25x                   & 36x                             \\ 
\cline{2-7}
                          & \multirow{2}{*}{ResNet9}  & \multirow{2}{*}{100}  & 800                   & 600                     & 8x                    & 6x                              \\ 
\cline{4-7}
                          &                           &                       & 3000                  & 3000                    & 30x                   & 30x                             \\ 
\cline{2-7}
                          & \multirow{2}{*}{ResNet11} & \multirow{2}{*}{100}  & \multirow{2}{*}{3500} & 600                     & \multirow{2}{*}{35x}  & 6x                              \\ 
\cline{5-5}\cline{7-7}
                          &                           &                       &                       & 3000                    &                       & 30x                             \\
\hline
\end{tabular}
\end{table}

\subsection{Complexity Reduction}
Deep ANNs struggle to meet the demand of extraordinary computational requirements. SNNs can mitigate this effort by enabling efficient event-based computations. To compare the computational complexity of these two cases, we first need to understand the operation principle of both. An ANN operation for inferring the category of a particular input requires a single feed-forward pass per image. For the same task, the network must be evaluated over a number of time-steps in the spiking domain. If regular hardware is used for both ANN and SNN, then it is evident that SNN will have computation complexity in the order of hundreds or thousands more compared to an ANN. However, there are specialized hardwares that account for the event-based neural operation and `computes only when required' for inference. SNNs can potentially exploit such alternative mechanisms of network operation and carry out an inference operation in the spiking domain much more efficiently than an ANN. Also, for deep SNNs, we have observed the increase in sparsity as the network depth increases. Hence, the benefits from event-based neuromorphic hardware are expected to increase as the network depth increases.\par

An estimate of the actual energy consumption of SNNs and comparison with ANNs is outside the scope of this work. However, we can gain some insight by quantifying the computational energy consumption for a synaptic operation and comparing the number of synaptic operations being performed in the ANN versus the SNN trained with our proposed algorithm and ANN-SNN converted network. We can estimate the number of synaptic operations per layer of a neural network from the structure for the convolutional and linear layers. In an ANN, a multiply-accumulate (MAC) computation is performed per synaptic operation. While a specialized SNN hardware would perform simply an accumulate computation (AC) per synaptic operation only if an incoming spike is received. Hence, the total number of AC operations in a SNN can be estimated by the layer-wise product and summation of the average neural spike count for a particular layer and the corresponding number of synaptic connections. We also have to multiply the \#time-steps with the \#AC operations to get total \#AC operation for one inference. For example, assume that there are $L$ layers each with $N_l$ neurons, $S_l$ synaptic connections and $a_l$ average spiking activity where $l$ is the layer number. Then, the total number of synaptic operations in a layer is $N_l \times S_l \times a_l$. The $N_l \times S_l$ is equal to the ANN (\#MAC) operations of a particular layer. Therefore, the total number of synaptic operations in a layer of an SNN becomes $\#MAC_l \times a_l$. The total number of AC operations required for an image inference is the sum of synaptic operations in all layers during the inference time-window. Hence, $\#AC/inference$=($\sum_{l}(\#MAC_l \times a_l)) \times \#timesteps$. This formula is used for estimating both ANN-SNN AC operations and SNN AC operations per image inference. On the other hand, the number of ANN (MAC) operation per inference becomes simply, $\#MAC/inference$=$\sum_{1}^{L}(\#MAC_l)$. Based on this concept, we estimated the total number of MAC operations for ANN, and the total number of AC operations for direct-spike trained SNN and ANN-SNN converted network, for VGG9, ResNet9 and ResNet11. The ratio of ANN-SNN converted networks' (scheme1-scheme2) AC operations to direct-spike trained SNN AC operations to ANN MAC operations is (28.18-25.60):3.61:1 for VGG9 while the ratio is (11.67-18.42):5.06:1 for the ResNet9 and (9.6-10.16):2.09:1 for ResNet11 (for maximum accuracy condition).

\begin{figure*}[h]
\centering
\includegraphics[width=\textwidth]{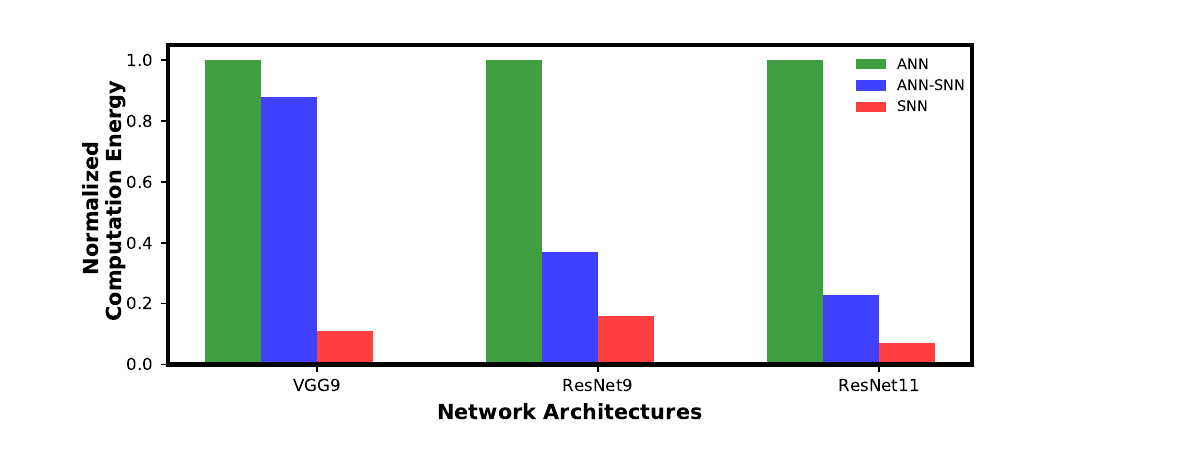}
\caption{Inference computation complexity comparison between ANN, ANN-SNN conversion and SNN trained with spike-based backpropagation. ANN computational complexity is considered as a baseline for normalization.}
\label{fig:fig8}
\end{figure*}

However, a MAC operation usually consumes an order of magnitude more energy than an AC operation. For instance, according to \cite{han2015learning}, a 32-bit floating point MAC operation consumes 4.6pJ and a 32-bit floating point AC operation consumes 0.9pJ in 45nm technology node. Hence, one synaptic operation in an ANN is equivalent to \texttildelow5 synaptic operations in a SNN. Moreover, 32-bit floating point computation can be replaced by fixed point computation using integer MAC and AC units without losing accuracy since the conversion is reported to be almost loss-less \cite{lin2016fixed}. A 32-bit integer MAC consumes roughly 3.2pJ while a 32-bit AC operation consumes only 0.1pJ in 45nm process technology. 
Considering this fact, our calculations demonstrate that the SNNs trained using the proposed method will be 7.81x\texttildelow7.10x and 8.87x more computationally energy-efficient for inference compared to the ANN-SNN converted networks and an ANN, respectively, for the VGG9 network architecture. 
We also gain 4.6x\texttildelow4.87x(2.31x\texttildelow3.64x) and 15.32x(6.32x) energy-efficiency, for the ResNet11(ResNet9) network, compared to the ANN-SNN converted networks and an ANN, respectively. The figure \ref{fig:fig8} shows the reduction in computation complexity for ANN-SNN conversions and SNN trained with the proposed methodology compared to ANNs.

It is worth noting here that as the sparsity of the spike signals increases with an increase in network depth in SNNs. Hence, the energy-efficiency is expected to increase almost exponentially in both ANN-SNN conversion network \cite{sengupta2018going} and SNN trained with proposed methodology compared to an ANN implementation. The depth of network is the key factor for achieving a significant increase in the energy efficiency for SNNs in neuromorphic hardware. However, the computational efficiency does not perfectly align with the overall efficiency since the dominant energy consumption can be the memory traffic on von-Neumann computing hardware. The dataflows in asynchronous SNNs are less predictable and more complicated. Hence, a detailed study is required to estimate the overall efficiency of SNNs accurately.

\begin{table}[h]
\centering
\caption{Iso-spike comparison for optimal condition. SNN time-steps corresponds to the latency to reach accuracy within \texttildelow1\% of maximum accuracy. ANN-SNN time-steps corresponds to the latency required for same number of spike/inference as SNN to occur. SNN and ANN-SNN accuracies are accuracies corresponding to respective latency.}
\label{IsoSpikeComp}
\begin{tabular}{|c|c|c|c|c|c|c|c|}
\hline
\multirow{2}{*}{\textbf{Dataset}}  & \multirow{2}{*}{\textbf{Model}} & \multicolumn{3}{c|}{\textbf{Time-steps}}          & \multicolumn{3}{c|}{\textbf{Accuracy (\%)}}            \\ \cline{3-8} 
                          &                        & \textbf{SNN} & \textbf{ANN-SNN} \cite{diehl2015fast} & \textbf{ANN-SNN} \cite{sengupta2018going} & \textbf{SNN} & \textbf{ANN-SNN} \cite{diehl2015fast} & \textbf{ANN-SNN} \cite{sengupta2018going}\\ \hline 
MNIST                     & LeNet                  & 20                   & 62      & 75      & 99.36                & 99.19   & 88.62   \\ \hline
\multirow{2}{*}{SVHN}     & VGG7                   & 30                   & 235     & 235     & 95.00                & 95.34   & 88.13   \\ \cline{2-8} 
                          & ResNet7                & 30                   & 200     & 200     & 95.06                & 95.63   & 95.48   \\ \hline
\multirow{3}{*}{CIFAR-10} & VGG9                   & 50                   & 228     & 260     & 89.33                & 69.53   & 61.08   \\ \cline{2-8} 
                          & ResNet9                & 50                   & 390     & 490     & 89.52                & 89.51   & 90.06   \\ \cline{2-8} 
                          & ResNet11               & 50                   & 307     & 280     & 90.24                & 82.75   & 73.82   \\ \hline
\end{tabular}%
\end{table}

\subsection{Iso-spike Comparison for Optimal Condition}
In section 4.3, we observe that SNNs trained with proposed method achieve significant speed-up in both max-accuracy and iso-accuracy condition. However, in section 4.2.2, we found that the proposed method is in some cases (in an iso-accuracy condition) not more efficient than ANN-SNN conversions in terms of \#spike/inference. The reason behind it is that an iso-accuracy condition may not be optimal for the SNNs trained with proposed method. In an iso-accuracy case, we have used max-accuracy latency (50 time-steps for MNIST and 100 time-steps for other networks) for direct-spike trained SNN, whereas most of the conversion networks used much less latency than the max-accuracy condition. In view of this, there is a need to determine the circumstances where our proposed method performs as well as or better than the SNN-ANN conversion methods on spike count, time steps, and accuracy. Consequently, in this section we analyze another interesting comparison.\par

In this analysis, we compare our proposed method and ANN-SNN conversion methods \cite{diehl2015fast,sengupta2018going} under the optimal condition at equal number of spikes. 
We define the `optimal \#time-steps' for SNNs trained with our proposed method as the \#time-steps required to reach within \texttildelow1\% of peak accuracy (when the accuracy starts to saturate). Based on this definition, we observed that the optimal \#time-steps for MNIST, SVHN, CIFAR10 networks are 20, 30, and 50, respectively. For this comparison, we recorded the achieved accuracy and \#spike/inference of the SNNs trained with our proposed method for the corresponding optimal \#time-steps. Then, we ran ANN-SNN networks for a length of time such that they use the similar number of spikes. In this iso-spike condition, we recorded the accuracy of the ANN-SNN networks (for both conversion methods) and the number of time-steps they require. The results are summarized in table \ref{IsoSpikeComp}. \par
For comparatively shallower networks such as LeNet, VGG7 (VGG type) and ResNet7, ResNet9 (Residual type), the ANN-SNN conversion networks achieve as good as or slightly better accuracy at iso-spike condition compared to the SNNs trained with our proposed method. However, these ANN-SNN conversion networks require 3x-10x higher latency for inference. 
On the other hand, for deeper networks such as VGG9 and ResNet11, the ANN-SNN conversion networks achieve significantly lower accuracy compared to SNNs trained with our proposed method even with much higher latency. This trend indicates that for deeper networks, SNNs trained with our proposed method will be more energy-efficient than the conversion networks under an iso-spike condition.

\section{Conclusion}
In this work, we propose a spike-based backpropagation training methodology for popular deep neural network architectures. This methodology enables deep SNNs to achieve competitive classification accuracies on standard image recognition tasks. Our experiments show the effectiveness of the proposed learning strategy on deeper SNNs (7-11 layer VGG and ResNet network architectures) by achieving the best classification accuracies in MNIST, SVHN and CIFAR-10 datasets among other networks trained with spike-based learning till date. The performance gap in terms of quality between ANN and SNN is substantially reduced by the application of our proposed methodology. Moreover, significant computational energy savings are expected when deep SNNs (trained with the proposed method) are employed on suitable neuromorphic hardware for the inference.

\section*{Acknowledgement}
This work was supported in part by C-BRIC, one of six centers in JUMP, a Semiconductor Research Corporation (SRC) program sponsored by DARPA, the National Science Foundation, Intel Corporation, the DoD Vannevar Bush Fellowship and the U.S. Army Research Laboratory and the U.K. Ministry of Defence under Agreement Number W911NF-16-3-0001. The views and conclusions contained in this document are those of the authors and should not be interpreted as representing the official policies, either expressed or implied, of the U.S. Army Research Laboratory, the U.S. Government, the U.K. Ministry of Defence or the U.K. Government. The U.S. and U.K. Governments are authorized to reproduce and distribute reprints for Government purposes notwithstanding any copyright notation hereon. We would like to thank Dr. Gerard (Rod) Rinkus for helpful comments.

\bibliography{template.bbl}

\begin{thebibliography}{10}

\bibitem{ankit2017resparc}
A.~Ankit, A.~Sengupta, P.~Panda, and K.~Roy.
\newblock Resparc: A reconfigurable and energy-efficient architecture with
  memristive crossbars for deep spiking neural networks.
\newblock In {\em Proceedings of the 54th Annual Design Automation Conference
  2017}, pages 1--6, 2017.

\bibitem{bellec2019solution}
G.~Bellec, F.~Scherr, A.~Subramoney, E.~Hajek, D.~Salaj, R.~Legenstein, and
  W.~Maass.
\newblock A solution to the learning dilemma for recurrent networks of spiking
  neurons.
\newblock {\em bioRxiv}, page 738385, 2019.

\bibitem{bengio2013estimating}
Y.~Bengio, N.~L{\'e}onard, and A.~Courville.
\newblock Estimating or propagating gradients through stochastic neurons for
  conditional computation.
\newblock {\em arXiv preprint arXiv:1308.3432}, 2013.

\bibitem{bohte2002error}
S.~M. Bohte, J.~N. Kok, and H.~La~Poutre.
\newblock Error-backpropagation in temporally encoded networks of spiking
  neurons.
\newblock {\em Neurocomputing}, 48(1-4):17--37, 2002.

\bibitem{brader2007learning}
J.~M. Brader, W.~Senn, and S.~Fusi.
\newblock Learning real-world stimuli in a neural network with spike-driven
  synaptic dynamics.
\newblock {\em Neural computation}, 19(11):2881--2912, 2007.

\bibitem{brette2005adaptive}
R.~Brette and W.~Gerstner.
\newblock Adaptive exponential integrate-and-fire model as an effective
  description of neuronal activity.
\newblock {\em Journal of neurophysiology}, 94(5):3637--3642, 2005.

\bibitem{cao2015spiking}
Y.~Cao, Y.~Chen, and D.~Khosla.
\newblock Spiking deep convolutional neural networks for energy-efficient
  object recognition.
\newblock {\em International Journal of Computer Vision}, 113(1):54--66, 2015.

\bibitem{davies2018loihi}
M.~Davies, N.~Srinivasa, T.-H. Lin, G.~Chinya, Y.~Cao, S.~H. Choday, G.~Dimou,
  P.~Joshi, N.~Imam, S.~Jain, et~al.
\newblock Loihi: A neuromorphic manycore processor with on-chip learning.
\newblock {\em IEEE Micro}, 38(1):82--99, 2018.

\bibitem{dayan2001theoretical}
P.~Dayan and L.~F. Abbott.
\newblock {\em Theoretical neuroscience}, volume 806.
\newblock Cambridge, MA: MIT Press, 2001.

\bibitem{diehl2015unsupervised}
P.~U. Diehl and M.~Cook.
\newblock Unsupervised learning of digit recognition using
  spike-timing-dependent plasticity.
\newblock {\em Frontiers in computational neuroscience}, 9:99, 2015.

\bibitem{diehl2015fast}
P.~U. Diehl, D.~Neil, J.~Binas, M.~Cook, S.-C. Liu, and M.~Pfeiffer.
\newblock Fast-classifying, high-accuracy spiking deep networks through weight
  and threshold balancing.
\newblock In {\em 2015 International Joint Conference on Neural Networks
  (IJCNN)}, pages 1--8. IEEE, 2015.

\bibitem{diehl2016conversion}
P.~U. Diehl, G.~Zarrella, A.~Cassidy, B.~U. Pedroni, and E.~Neftci.
\newblock Conversion of artificial recurrent neural networks to spiking neural
  networks for low-power neuromorphic hardware.
\newblock In {\em Rebooting Computing (ICRC), IEEE International Conference
  on}, pages 1--8. IEEE, 2016.

\bibitem{esser2016convolutional}
S.~Esser, P.~Merolla, J.~Arthur, A.~Cassidy, R.~Appuswamy, A.~Andreopoulos,
  D.~Berg, J.~McKinstry, T.~Melano, D.~Barch, et~al.
\newblock Convolutional networks for fast, energy-efficient neuromorphic
  computing. 2016.
\newblock {\em Preprint on ArXiv. http://arxiv. org/abs/1603.08270. Accessed},
  27, 2016.

\bibitem{furber2013overview}
S.~B. Furber, D.~R. Lester, L.~A. Plana, J.~D. Garside, E.~Painkras, S.~Temple,
  and A.~D. Brown.
\newblock Overview of the spinnaker system architecture.
\newblock {\em IEEE Transactions on Computers}, 62(12):2454--2467, 2013.

\bibitem{han2015learning}
S.~Han, J.~Pool, J.~Tran, and W.~Dally.
\newblock Learning both weights and connections for efficient neural network.
\newblock In {\em Advances in neural information processing systems}, pages
  1135--1143, 2015.

\bibitem{he2015delving}
K.~He, X.~Zhang, S.~Ren, and J.~Sun.
\newblock Delving deep into rectifiers: Surpassing human-level performance on
  imagenet classification.
\newblock In {\em Proceedings of the IEEE international conference on computer
  vision}, pages 1026--1034, 2015.

\bibitem{he2016deep}
K.~He, X.~Zhang, S.~Ren, and J.~Sun.
\newblock Deep residual learning for image recognition.
\newblock In {\em Proceedings of the IEEE conference on computer vision and
  pattern recognition}, pages 770--778, 2016.

\bibitem{hodgkin1952currents}
A.~L. Hodgkin and A.~F. Huxley.
\newblock Currents carried by sodium and potassium ions through the membrane of
  the giant axon of loligo.
\newblock {\em The Journal of physiology}, 116(4):449--472, 1952.

\bibitem{huh2018gradient}
D.~Huh and T.~J. Sejnowski.
\newblock Gradient descent for spiking neural networks.
\newblock In {\em Advances in Neural Information Processing Systems}, pages
  1440--1450, 2018.

\bibitem{hunsberger2015spiking}
E.~Hunsberger and C.~Eliasmith.
\newblock Spiking deep networks with lif neurons.
\newblock {\em arXiv preprint arXiv:1510.08829}, 2015.

\bibitem{ioffe2015batch}
S.~Ioffe and C.~Szegedy.
\newblock Batch normalization: Accelerating deep network training by reducing
  internal covariate shift.
\newblock {\em arXiv preprint arXiv:1502.03167}, 2015.

\bibitem{izhikevich2003simple}
E.~M. Izhikevich.
\newblock Simple model of spiking neurons.
\newblock {\em IEEE Transactions on neural networks}, 14(6):1569--1572, 2003.

\bibitem{jin2018hybrid}
Y.~Jin, P.~Li, and W.~Zhang.
\newblock Hybrid macro/micro level backpropagation for training deep spiking
  neural networks.
\newblock {\em arXiv preprint arXiv:1805.07866}, 2018.

\bibitem{kappel2015network}
D.~Kappel, S.~Habenschuss, R.~Legenstein, and W.~Maass.
\newblock Network plasticity as bayesian inference.
\newblock {\em PLoS computational biology}, 11(11):e1004485, 2015.

\bibitem{kappel2018dynamic}
D.~Kappel, R.~Legenstein, S.~Habenschuss, M.~Hsieh, and W.~Maass.
\newblock A dynamic connectome supports the emergence of stable computational
  function of neural circuits through reward-based learning.
\newblock {\em Eneuro}, 5(2), 2018.

\bibitem{kheradpisheh2016stdp}
S.~R. Kheradpisheh, M.~Ganjtabesh, S.~J. Thorpe, and T.~Masquelier.
\newblock Stdp-based spiking deep neural networks for object recognition.
\newblock {\em arXiv preprint arXiv:1611.01421}, 2016.

\bibitem{kingma2014adam}
D.~P. Kingma and J.~Ba.
\newblock Adam: A method for stochastic optimization.
\newblock {\em arXiv preprint arXiv:1412.6980}, 2014.

\bibitem{krizhevsky2009learning}
A.~Krizhevsky and G.~Hinton.
\newblock Learning multiple layers of features from tiny images.
\newblock Technical report, Citeseer, 2009.

\bibitem{krizhevsky2012imagenet}
A.~Krizhevsky, I.~Sutskever, and G.~E. Hinton.
\newblock Imagenet classification with deep convolutional neural networks.
\newblock In {\em Advances in neural information processing systems}, pages
  1097--1105, 2012.

\bibitem{lecun1998gradient}
Y.~LeCun, L.~Bottou, Y.~Bengio, and P.~Haffner.
\newblock Gradient-based learning applied to document recognition.
\newblock {\em Proceedings of the IEEE}, 86(11):2278--2324, 1998.

\bibitem{lee2018training}
C.~Lee, P.~Panda, G.~Srinivasan, and K.~Roy.
\newblock Training deep spiking convolutional neural networks with stdp-based
  unsupervised pre-training followed by supervised fine-tuning.
\newblock {\em Frontiers in Neuroscience}, 12:435, 2018.

\bibitem{lee2018deep}
C.~Lee, G.~Srinivasan, P.~Panda, and K.~Roy.
\newblock Deep spiking convolutional neural network trained with unsupervised
  spike timing dependent plasticity.
\newblock {\em IEEE Transactions on Cognitive and Developmental Systems}, 2018.

\bibitem{lee2016training}
J.~H. Lee, T.~Delbruck, and M.~Pfeiffer.
\newblock Training deep spiking neural networks using backpropagation.
\newblock {\em Frontiers in neuroscience}, 10:508, 2016.

\bibitem{lichtsteiner2008128}
P.~Lichtsteiner, C.~Posch, and T.~Delbruck.
\newblock A 128$\times$128 120 db 15$\mu$s latency asynchronous temporal
  contrast vision sensor.
\newblock {\em IEEE journal of solid-state circuits}, 43(2):566--576, 2008.

\bibitem{lillicrap2019backpropagation}
T.~P. Lillicrap and A.~Santoro.
\newblock Backpropagation through time and the brain.
\newblock {\em Current opinion in neurobiology}, 55:82--89, 2019.

\bibitem{lin2016fixed}
D.~Lin, S.~Talathi, and S.~Annapureddy.
\newblock Fixed point quantization of deep convolutional networks.
\newblock In {\em International Conference on Machine Learning}, pages
  2849--2858, 2016.

\bibitem{maass1997networks}
W.~Maass.
\newblock Networks of spiking neurons: the third generation of neural network
  models.
\newblock {\em Neural networks}, 10(9):1659--1671, 1997.

\bibitem{merolla2014million}
P.~A. Merolla, J.~V. Arthur, R.~Alvarez-Icaza, A.~S. Cassidy, J.~Sawada,
  F.~Akopyan, B.~L. Jackson, N.~Imam, C.~Guo, Y.~Nakamura, et~al.
\newblock A million spiking-neuron integrated circuit with a scalable
  communication network and interface.
\newblock {\em Science}, 345(6197):668--673, 2014.

\bibitem{mostafa2017supervised}
H.~Mostafa.
\newblock Supervised learning based on temporal coding in spiking neural
  networks.
\newblock {\em IEEE transactions on neural networks and learning systems},
  2017.

\bibitem{neftci2017event}
E.~O. Neftci, C.~Augustine, S.~Paul, and G.~Detorakis.
\newblock Event-driven random back-propagation: Enabling neuromorphic deep
  learning machines.
\newblock {\em Frontiers in neuroscience}, 11:324, 2017.

\bibitem{neftci2015unsupervised}
E.~O. Neftci, B.~U. Pedroni, S.~Joshi, M.~Al-Shedivat, and G.~Cauwenberghs.
\newblock Unsupervised learning in synaptic sampling machines.
\newblock {\em arXiv preprint arXiv:1511.04484}, 2015.

\bibitem{netzer2011reading}
Y.~Netzer, T.~Wang, A.~Coates, A.~Bissacco, B.~Wu, and A.~Y. Ng.
\newblock Reading digits in natural images with unsupervised feature learning.
\newblock In {\em NIPS workshop on deep learning and unsupervised feature
  learning}, volume 2011, page~5, 2011.

\bibitem{orchard2015converting}
G.~Orchard, A.~Jayawant, G.~K. Cohen, and N.~Thakor.
\newblock Converting static image datasets to spiking neuromorphic datasets
  using saccades.
\newblock {\em Frontiers in neuroscience}, 9:437, 2015.

\bibitem{panda2016unsupervised}
P.~Panda and K.~Roy.
\newblock Unsupervised regenerative learning of hierarchical features in
  spiking deep networks for object recognition.
\newblock In {\em Neural Networks (IJCNN), 2016 International Joint Conference
  on}, pages 299--306. IEEE, 2016.

\bibitem{paszke2017automatic}
A.~Paszke, S.~Gross, S.~Chintala, G.~Chanan, E.~Yang, Z.~DeVito, Z.~Lin,
  A.~Desmaison, L.~Antiga, and A.~Lerer.
\newblock Automatic differentiation in pytorch.
\newblock 2017.

\bibitem{rueckauer2017conversion}
B.~Rueckauer, I.-A. Lungu, Y.~Hu, M.~Pfeiffer, and S.-C. Liu.
\newblock Conversion of continuous-valued deep networks to efficient
  event-driven networks for image classification.
\newblock {\em Frontiers in neuroscience}, 11:682, 2017.

\bibitem{rumelhart1985learning}
D.~E. Rumelhart, G.~E. Hinton, and R.~J. Williams.
\newblock Learning internal representations by error propagation.
\newblock Technical report, California Univ San Diego La Jolla Inst for
  Cognitive Science, 1985.

\bibitem{sarwar2018energy}
S.~S. Sarwar, G.~Srinivasan, B.~Han, P.~Wijesinghe, A.~Jaiswal, P.~Panda,
  A.~Raghunathan, and K.~Roy.
\newblock Energy efficient neural computing: A study of cross-layer
  approximations.
\newblock {\em IEEE Journal on Emerging and Selected Topics in Circuits and
  Systems}, 2018.

\bibitem{sengupta2018going}
A.~Sengupta, Y.~Ye, R.~Wang, C.~Liu, and K.~Roy.
\newblock Going deeper in spiking neural networks: Vgg and residual
  architectures.
\newblock {\em Frontiers in neuroscience}, 13, 2019.

\bibitem{shrestha2018slayer}
S.~B. Shrestha and G.~Orchard.
\newblock Slayer: Spike layer error reassignment in time.
\newblock In {\em Advances in Neural Information Processing Systems}, pages
  1412--1421, 2018.

\bibitem{silver2016mastering}
D.~Silver, A.~Huang, C.~J. Maddison, A.~Guez, L.~Sifre, G.~Van Den~Driessche,
  J.~Schrittwieser, I.~Antonoglou, V.~Panneershelvam, M.~Lanctot, et~al.
\newblock Mastering the game of go with deep neural networks and tree search.
\newblock {\em nature}, 529(7587):484, 2016.

\bibitem{simonyan2014very}
K.~Simonyan and A.~Zisserman.
\newblock Very deep convolutional networks for large-scale image recognition.
\newblock {\em arXiv preprint arXiv:1409.1556}, 2014.

\bibitem{srinivasan2018spilinc}
G.~Srinivasan, P.~Panda, and K.~Roy.
\newblock Spilinc: Spiking liquid-ensemble computing for unsupervised speech
  and image recognition.
\newblock {\em Frontiers in Neuroscience}, 12:524, 2018.

\bibitem{srinivasan2018stdp}
G.~Srinivasan, P.~Panda, and K.~Roy.
\newblock Stdp-based unsupervised feature learning using convolution-over-time
  in spiking neural networks for energy-efficient neuromorphic computing.
\newblock {\em ACM Journal on Emerging Technologies in Computing Systems
  (JETC)}, 14(4):44, 2018.

\bibitem{srivastava2014dropout}
N.~Srivastava, G.~Hinton, A.~Krizhevsky, I.~Sutskever, and R.~Salakhutdinov.
\newblock Dropout: a simple way to prevent neural networks from overfitting.
\newblock {\em The Journal of Machine Learning Research}, 15(1):1929--1958,
  2014.

\bibitem{tavanaei2016bio}
A.~Tavanaei and A.~S. Maida.
\newblock Bio-inspired spiking convolutional neural network using layer-wise
  sparse coding and stdp learning.
\newblock {\em arXiv preprint arXiv:1611.03000}, 2016.

\bibitem{tavanaei2017multi}
A.~Tavanaei and A.~S. Maida.
\newblock Multi-layer unsupervised learning in a spiking convolutional neural
  network.
\newblock In {\em Neural Networks (IJCNN), 2017 International Joint Conference
  on}, pages 2023--2030. IEEE, 2017.

\bibitem{werbos1990backpropagation}
P.~J. Werbos et~al.
\newblock Backpropagation through time: what it does and how to do it.
\newblock {\em Proceedings of the IEEE}, 78(10):1550--1560, 1990.

\bibitem{wu2018direct}
Y.~Wu, L.~Deng, G.~Li, J.~Zhu, and L.~Shi.
\newblock Direct training for spiking neural networks: Faster, larger, better.
\newblock {\em arXiv preprint arXiv:1809.05793}, 2018.

\bibitem{wu2018spatio}
Y.~Wu, L.~Deng, G.~Li, J.~Zhu, and L.~Shi.
\newblock Spatio-temporal backpropagation for training high-performance spiking
  neural networks.
\newblock {\em Frontiers in neuroscience}, 12, 2018.

\bibitem{zhao2015feedforward}
B.~Zhao, R.~Ding, S.~Chen, B.~Linares-Barranco, and H.~Tang.
\newblock Feedforward categorization on aer motion events using cortex-like
  features in a spiking neural network.
\newblock {\em IEEE transactions on neural networks and learning systems},
  26(9):1963--1978, 2015.

\end{thebibliography}

\end{document}